\documentclass[11pt]{article}

\usepackage[final]{acl}

\usepackage{times}
\usepackage{latexsym}
\usepackage{enumitem}
\usepackage[T1]{fontenc}
\usepackage{placeins}

\usepackage[utf8]{inputenc}
\usepackage[svgnames]{xcolor}

\usepackage{microtype}

\usepackage{inconsolata}

\usepackage{xspace}
\usepackage{graphicx}
\usepackage{booktabs}

\newcommand{\etc}{etc\@\xspace}
\newcommand{\eg}{e.g.\@\xspace}
\usepackage[capitalize]{cleveref}
\crefname{section}{Sec.}{Secs.}
\Crefname{section}{Sec.}{Secs.}
\crefname{appendix}{Appendix}{Appendices}
\Crefname{appendix}{Appendix}{Appendices}
\crefname{table}{Tab.}{Tabs.}
\Crefname{table}{Tab.}{Tabs.}
\crefname{figure}{Fig.}{Figs.}
\Crefname{figure}{Fig.}{Figs.}

\newcommand{\inlinelogo}[1]{%
  \raisebox{-0.2ex}{\includegraphics[height=1em]{#1}}%
}

\title{\ourbenchmark: Benchmarking Computer-Use Agents on Vast-Horizon Repetitive Tasks}

\author{
 \textbf{Jing Wu\textsuperscript{1} \footnotemark[1]},
 \textbf{Wenjie Ai \textsuperscript{4}},
 \textbf{Daphne Barretto\textsuperscript{2}},
 \textbf{Yiye Chen\textsuperscript{3} \footnotemark[1]}, 
 \textbf{Qingyu Chen\textsuperscript{5}}\\
 \textbf{Yuhang He\textsuperscript{2}},
 \textbf{Pranit Chawla\textsuperscript{2} \footnotemark[2]}
 \textbf{Nicholas Gydé\textsuperscript{2}},
 \textbf{Yanan Jian\textsuperscript{2} \footnotemark[2]},
 \textbf{Vibhav Vineet\textsuperscript{2}
}
\\
 \textsuperscript{1}University of Oxford,
 \textsuperscript{2}Microsoft,
 \textsuperscript{3}Georgia Institute of Technology \\
 \textsuperscript{4}University of Surrey,
 \textsuperscript{5}University of Washington
\\
 \small{
   \textbf{Correspondence:} \href{mailto:jing.wu@eng.ox.ac.uk}{jing.wu@eng.ox.ac.uk}
 }
}

\begin{document}

\twocolumn[{%
\renewcommand\twocolumn[1][]{#1}%
\maketitle
\begin{center}
    \centering
\includegraphics[width=.95\textwidth]{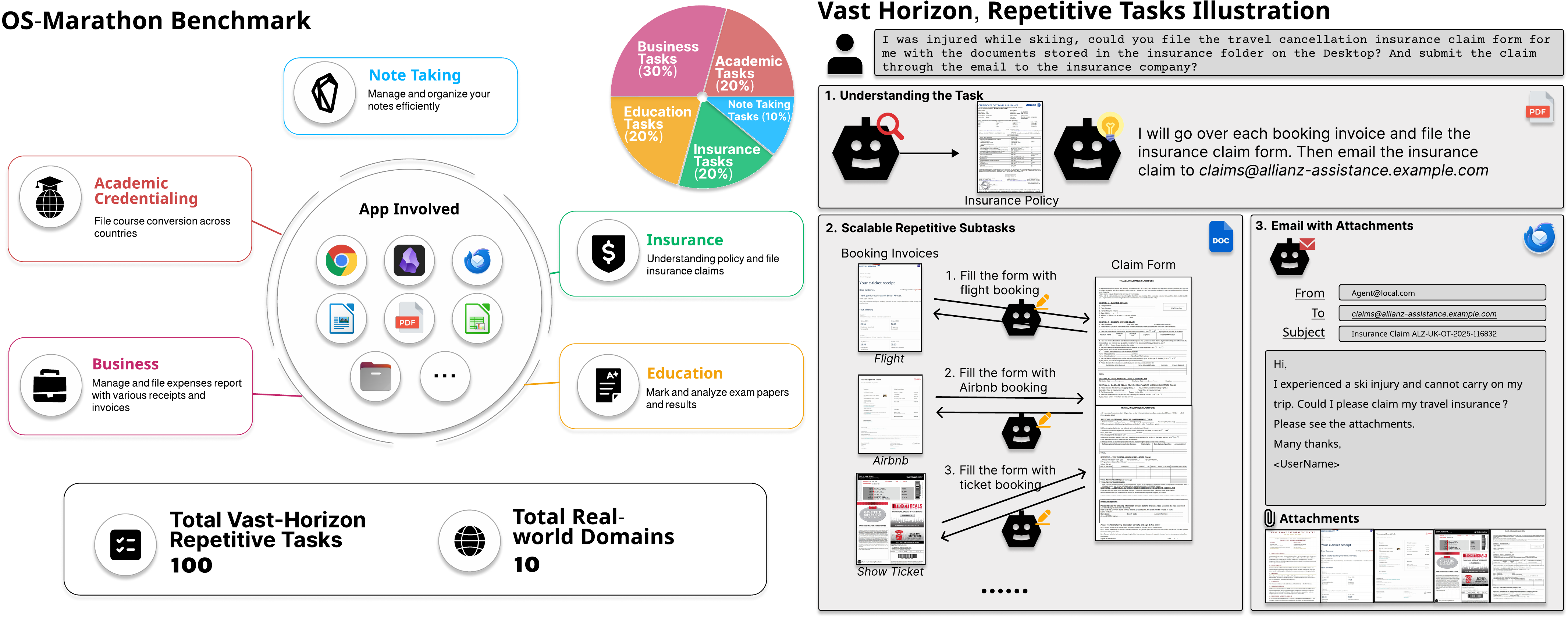}
    \captionof{figure}{Overview of \textbf{\ourbenchmark}. \ourbenchmark contains \tasknum vast-horizon, repetitive tasks, spanning \scenarionum scenarios, \domainnum domains, covering diverse daily use cases (Left). Successfully completing such a task requires three agent capabilities (Right): \emph{(i)} interpreting a high-level user intention and understanding the task instruction, \emph{(ii)} executing those subtasks iteratively and consistently, and \emph{(iii)} performing the closing actions that wrap up the task once all subtasks are done.}
\label{fig:teaser}
\end{center}
}]

\footnotetext[1]{Work completed during an internship at Microsoft.}
\footnotetext[2]{Work completed during employment at Microsoft.}

\begin{abstract}
\vspace{-3mm}
Vast-horizon, repetitive workflows are common in daily routines, e.g., processing expense reports from a stack of receipts, organising a collection of PDF annotations into structured notes, and are tedious for humans, with execution length scaling with the volume of data to process.
They are, however,  well-suited to autonomous agents, since their structured, recurring sub-workflows follow a logic that can be systematically learned.
Yet no existing benchmark evaluates agents specifically on vast-horizon, repetitive capabilities, leaving the problem largely underexplored.
To bridge this gap, we establish \textbf{\ourbenchmark}, comprising \tasknum vast-horizon repetitive tasks across \scenarionum scenarios and \domainnum domains, on which our evaluation reveals that leading state-of-the-art (SOTA) CUAs struggle substantially.
We further find that introducing a task orchestrator to decompose the workflow into per-instance subtasks fails to mitigate the challenge: errors accumulate and propagate across solver agents, indicating that naive decomposition is insufficient for these tasks.
We then explore a cost-friendly personalisation strategy, i.e. \ourmethod, that adapts general agents to such tasks from a single human demonstration of the recurring sub-workflow logic. 
Extensive experiments show both the real-world challenge of vast-horizon, repetitive tasks and the improvement provided by human demonstration in this setting.
Project website: \url{https://os-marathon.github.io/}. 
\end{abstract}

\section{Introduction}

Vast-horizon, repetitive workflows are prevalent in both professional and personal contexts. 
It generally involves processing a collection of data instances, requiring executing a series of similar sub-workflows to handle each data instance iteratively. 
For example, in professional settings, an expense-reporting workflow requires entering a set of receipts into a single report. Analogously, in personal settings, individuals develop their own repetitive routines shaped by their specific role, tools, and habits, e.g., a researcher reorganising a set of PDF paper-reading annotations into a personal preferred note format, or a teacher marking and analysing student exam papers of a class in a fixed format.
Such workflows are tedious for humans due to their extended duration and scalable repetitive nature.

\begin{figure}[t]
  \centering
  \includegraphics[width=1\linewidth]{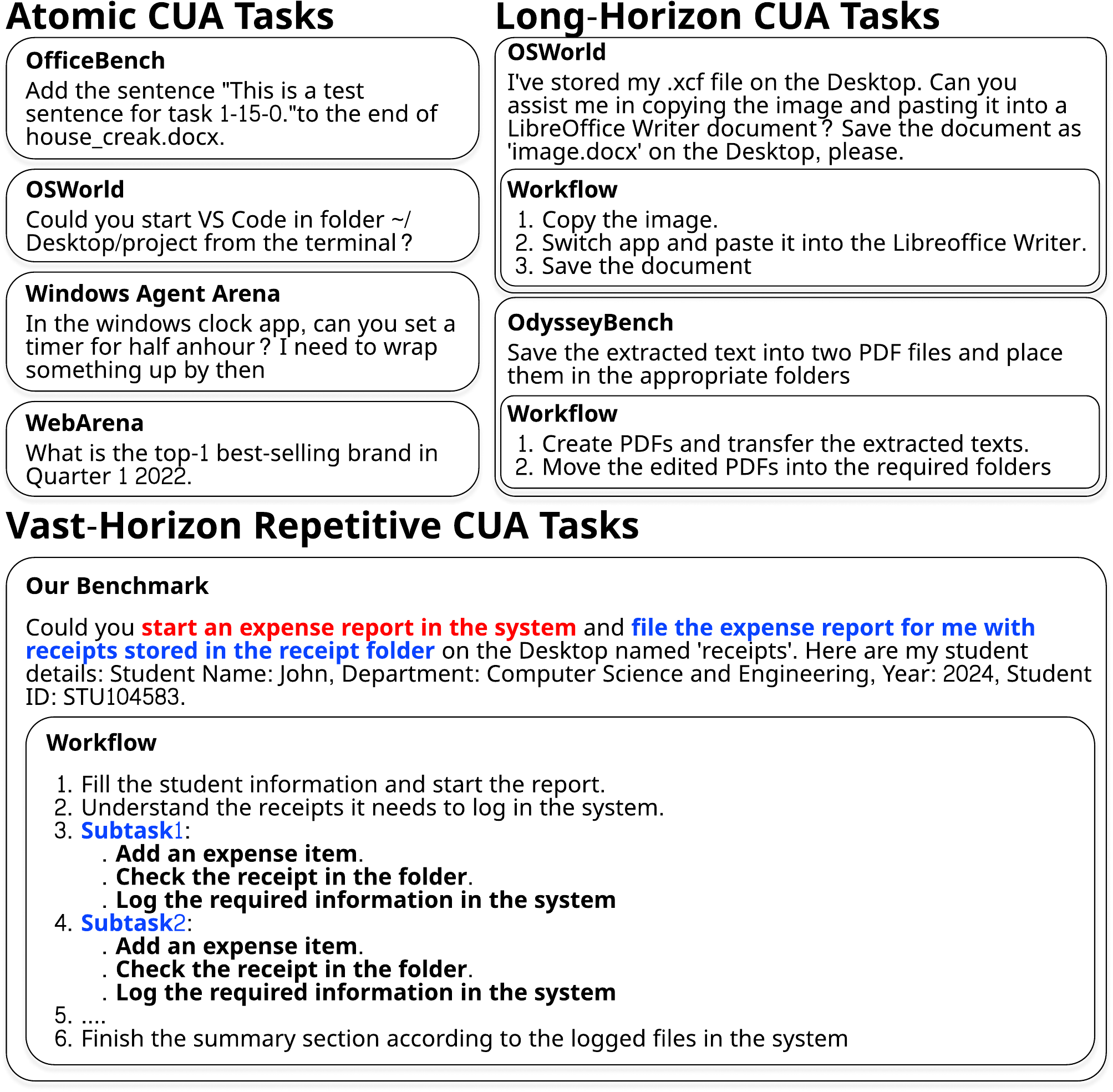}
  \caption{\textbf{Task comparison.} Atomic tasks are self-contained and require no long context. Long-horizon tasks chain atomic tasks into non-repetitive workflows whose instructions enumerate them explicitly. Vast-horizon, repetitive tasks scale with the number of data instances and are specified by high-level instructions that the agent must decompose itself. 
  }
  \label{fig:benchmark-task-type-comparison}
  \vspace{-6mm}
\end{figure}

Their repetitive structure, however, makes them a natural target for autonomous agents to alleviate human effort. 
Despite this potential, no existing benchmarks adequately stress-tests CUAs on real-world, vast-horizon, repetitive tasks. 
As illustrated in \cref{fig:benchmark-task-type-comparison}, existing benchmarks~\cite{wang2024officebench,bonatti2024WAA,zhou2023webarena,OSWorld,Wang2025OdysseyBenchEL} concentrate on two task types: \textbf{self-contained atomic tasks}, independent of prior interactions or accumulated context, and \textbf{non-repetitive long-horizon workflows} whose instructions may explicitly enumerate the atomic tasks to perform (e.g., \emph{save the files and place them in separate folders}).
Neither captures the two challenges unique to real-world vast-horizon, repetitive tasks: \textbf{a scalable horizon} and \textbf{the need to decompose a high-level user intention}.
The task horizon scales with the number of data instances, and the instruction itself also cannot be enumerated step-by-step in advance; for example, \emph{file an expense report from the receipts on the desktop} requires the agent to infer the report's internal structure and decompose the work into subtasks on its own. 
As a result, this task type lacks both a formal definition and a standardised evaluation framework for CUA capabilities.


To address this challenge, we formally define the problem and introduce \textbf{\ourbenchmark}, a benchmark designed specifically to evaluate CUA performance on vast-horizon, repetitive workflows. 
\ourbenchmark comprises \tasknum tasks, spanning \scenarionum scenarios, \domainnum domains, covering both professional and personal use cases (\cref{fig:teaser} (Left)).
Each \ourbenchmark task tests three agent capabilities, illustrated in \cref{fig:teaser} (Right). 
First, the agent must interpret a high-level user intention and plan how to decompose the work into per-instance subtasks.
Second, the agent needs to execute those subtasks repeatedly and consistently across all data instances; this is the stage that scales with the number of instances and drives the vast horizon of \ourbenchmark tasks.
Third, it must perform the closing actions that fall outside any individual subtask, e.g. completing summary sections or submitting the final output.


We evaluate leading CUAs on \ourbenchmark and analyse their performance and bottlenecks at multiple granularity levels, \eg atomic task execution, subtask-internal logic, and vast-horizon consistency \etc, providing a fine-grained understanding of how and where each agent fails.
The evaluation reveals that even leading CUAs fall behind human performance on \ourbenchmark, 
leaving a substantial gap between general-purpose agents and the user-specific, vast-horizon, repetitive workflows they would need to execute in practice.

To bridge the performance gap, we explore two routes. 
We first investigate using a task orchestrator that decomposes the workflow into per-instance subtasks to help agents on this task, but find it counterproductive: \textit{an error made by an earlier solver agent propagates forward and affects subsequent ones}, showing that this task is not solvable via simple decompositions. 
The second route exploits the repetitive structure of \ourbenchmark and suggests a cost-friendly personalisation strategy \ourmethod: since every per-instance subtask shares a similar execution logic, a single human demonstration on a small number of subtasks could, in principle, teach the agent the execution logic of the remaining ones. 
We structure the demonstration as a causal graph to support the vast horizons of \ourbenchmark tasks, providing the first reference point for personalising general CUAs on vast-horizon, repetitive tasks.

The contributions of this work are as follows: \\
\vspace{-5mm}
\begin{itemize}[nosep]
    \item We formally define vast-horizon, repetitive CUA tasks, identifying two properties that distinguish them from prior benchmarks: a scalable repetitive sub-workflow structure that drives the task length, and high-level user intentions that the agent must decompose on its own.
    \item We introduce \textbf{\ourbenchmark}, the first diverse benchmark tailored to this setting, comprising \tasknum tasks across \scenarionum scenarios, \domainnum domains, and \envnum execution environments. 
    \item We evaluate leading CUAs on \ourbenchmark across multiple granularity levels and benchmark them against human performance, demonstrating the real-world challenges that \ourbenchmark poses.
    \item We explore agent personalisation from a single human demonstration as a first step toward bridging the gap between general-purpose agents and the user-specific, vast-horizon, repetitive workflows. 
\end{itemize}

\section{Related Works}


\noindent\textbf{Benchmarks. } 
CUA benchmarks fall into two broad families. \textbf{Component-specific} benchmarks target fine-grained capabilities in isolation, such as UI grounding accuracy~\cite{li2025screenspotpro, liu2024visualwebbench} and trajectory-level behaviour~\cite{wang2025opencua}. These benchmarks are useful for diagnosing low-level skills but do not measure how well an agent composes those skills into a complete workflow.
\textbf{End-to-end} benchmarks measure task success across full executions and span desktop~\cite{OSWorld, abhyankar2025osworldhuman}, web~\cite{deng2023mind2web}, and mobile~\cite{rawles2024androidworld, lu2024guiOdyssey, liu2025learnact} domains, where most tasks are short-horizon. 
OdysseyBench~\cite{Wang2025OdysseyBenchEL,xu2026odysseyarena} introduces longer office workflows, pushing the field toward more realistic horizons. 
However, it does not capture vast-horizon, repetitive capabilities, where the horizon scales with the number of data instances.

\noindent\textbf{Computer-Use Agents. }
Existing computer-use agents follow two primary paradigms: \textbf{end-to-end specialised models} and \textbf{agentic frameworks}.
End-to-end systems~\cite{liu2024autoglm, fu2025DeepMiner-Mano, wang2025opencua} train a monolithic model to manage the entire perception-reasoning-action loop, encompassing planning, grounding, execution, and reflection.
Notable examples include AGUVIS~\cite{xu2024aguvis-bak} for unified vision-based control, UI-TARS~\cite{Qin2025UI-tars} for human-like keyboard and mouse input, and OpenAI's Operator~\cite{openaicua}, which combines GPT-4o with RL-based GUI control.
In contrast, agentic frameworks~\cite{xie2025jedi, yang2025gta1, yu2025aworld, guo2025agenticlybic-maestro, cheng2024seeclick, wu2024os-atlas} decompose the workflow into modular components (planning, grounding, execution) managed by an orchestrator.
For instance, AgentS~\cite{Agent-S, Agent-S2} sequences these modules explicitly. 
CoAct~\cite{hou2024coact} coordinates distinct code and GUI agents for CUA tasks.
Despite this progress, current agents struggle with vast-horizon, repetitive tasks: they fail to grasp the logic of individual subtasks and to execute that logic consistently across the vast horizon.


\noindent\textbf{Agent Personalisation. }
Adapting agents to individual users' workflows and preferences is an emerging challenge. 
Existing approaches fall into two categories: \textbf{memory-based} methods, which accumulate interaction experiences at inference time, and \textbf{demonstration-based} methods, which leverage explicit human demonstrations as guidance. Memory-based approaches include ICAL~\cite{sarch2024ical}, which distils interaction experiences into reusable memories for VLM agents, and PerPilot~\cite{wang2025PerPilot}, which personalises mobile agents through memory retrieval and reasoning-based exploration. 
Demonstration-based approaches include AdaptAgent~\cite{verma2024adaptagent} and LearnAct~\cite{liu2025learnact}, which use few-shot demonstrations to aid web and mobile agent generalisation; ShowUI-Aloha~\cite{showui_aloha} and Synapse~\cite{zheng2023synapse}, which inject full demonstration trajectories as in-context guidance for computer control; and IFR Agent~\cite{wu2025quickuptake}, which leverages demonstrations to interpret user intent in mobile settings.
However, these methods target short-horizon tasks where the task demonstration fits within the model's context window.  
For vast-horizon, repetitive workflows, this assumption breaks: injecting the entire task trajectory exceeds context limits, while retrieving complete demonstrations operates at the wrong granularity for tasks where the agent needs step-level guidance within a single extended execution.

\section{\ourbenchmark Benchmark}


    

\subsection{Task Definition}
\label{sec:task-definition}
\textbf{Preliminaries. }
We formulate a computer-use agent task as a Partially Observable Markov Decision Process (POMDP) ($\mathcal{S}$, $\mathcal{O}$, $\mathcal{A}$, $\mathcal{T}$, $\mathcal{R}$).
Following OSWorld \cite{OSWorld}, the state space $\mathcal{S}$ captures the underlying desktop environment, and the observation space $\mathcal{O}$ collects the multimodal computer signals, including a screenshot, a natural-language instruction, an accessibility (a11y) tree, or any combination thereof, depending on available modalities. 
The action space $\mathcal{A}$ contains all actions executable by the agent.
The transition function $\mathcal{T}: \mathcal{S} \times \mathcal{A} \rightarrow \mathcal{S}$ governs how the desktop state evolves under each action. 
The reward function $\mathcal{R}: \mathcal{S} \times \mathcal{A} \rightarrow [0,1]$ offers a sparse signal, assigning a reward of 1 only at the final step if and only if the task is completed.

\noindent\textbf{Vast-horizon, repetitive tasks.}
We specialise this POMDP to tasks that consist of sequentially processing a set of $N$ independent data instances $\{ d_i\}_{i=1}^{N}$, where each $d_i$ is a distinct item (e.g., one receipt to be logged to the expense system, or one student's exam to be graded) and $N$ denotes the total number of items to be processed. 
The complete workflow $\mathbf{M}$ decomposes into a pre-subtask phase $\mathbf{m}_{\text{pre}}$, $N$ repetitive subtask phases $\{ \mathbf{m}_{i} \}_{i=1}^{N}$, and a post-subtask phase $\mathbf{m}_{\text{post}}$:
\begin{align*}
    \textbf{M} = \{ \mathbf{m}_{\text{pre}}, \mathbf{m}_1, \mathbf{m}_2, ..., \mathbf{m}_N, \mathbf{m}_{\text{post}} \}.
\end{align*}
Each $\mathbf{m}_i$ is a specific POMDP workflow with $l_i$ POMDP steps used to process a subtask, corresponding to the data instance $d_i$ (e.g., logging one receipt into the form).
\begin{equation*}
\begin{split}
    \mathbf{m}_i = \{ & (S_{i,j}, O_{i,j}, A_{i,j}, T_{i,j}, R_{i,j}) | j \in [1,l_i] \}.
\end{split}
\end{equation*}
where the actions $A_{i,j}$ is conditioned on the specific information contained in $d_i$ (e.g., $\texttt{ExtractInfo}(d_i)$ reads a vendor name from a receipt).
In our formulation, the high-level execution logic of $\mathbf{m}_i$ is shared across $i$, while only the data $d_i$ varies.
The pre- and post-subtask phases $\mathbf{m}_{\text{pre}}$ and $\mathbf{m}_{\text{post}}$ are two one-time workflows: $\mathbf{m}_{\text{pre}}$ prepares the environment before any data is processed (e.g., filling in shared meta-information on the expense form, such as the employee's name and ID) and $\mathbf{m}_{\text{post}}$ runs after all data instances are processed (e.g., emailing the completed insurance claim to the insurance company once every booking has been entered in the form). 
They are standard POMDP trajectories not directly conditioned on $d_i$, yet they share the execution environment with the per-instance subtasks $\mathbf{m}_i$ and are therefore indirectly shaped by the data through the environment state.
The reward function assigns a reward of 1 only upon the completion of the post-subtask phase, i.e., $R_{\text{post}} = 1$.
\begin{figure}[t]
  \centering
  \includegraphics[width=1\linewidth]{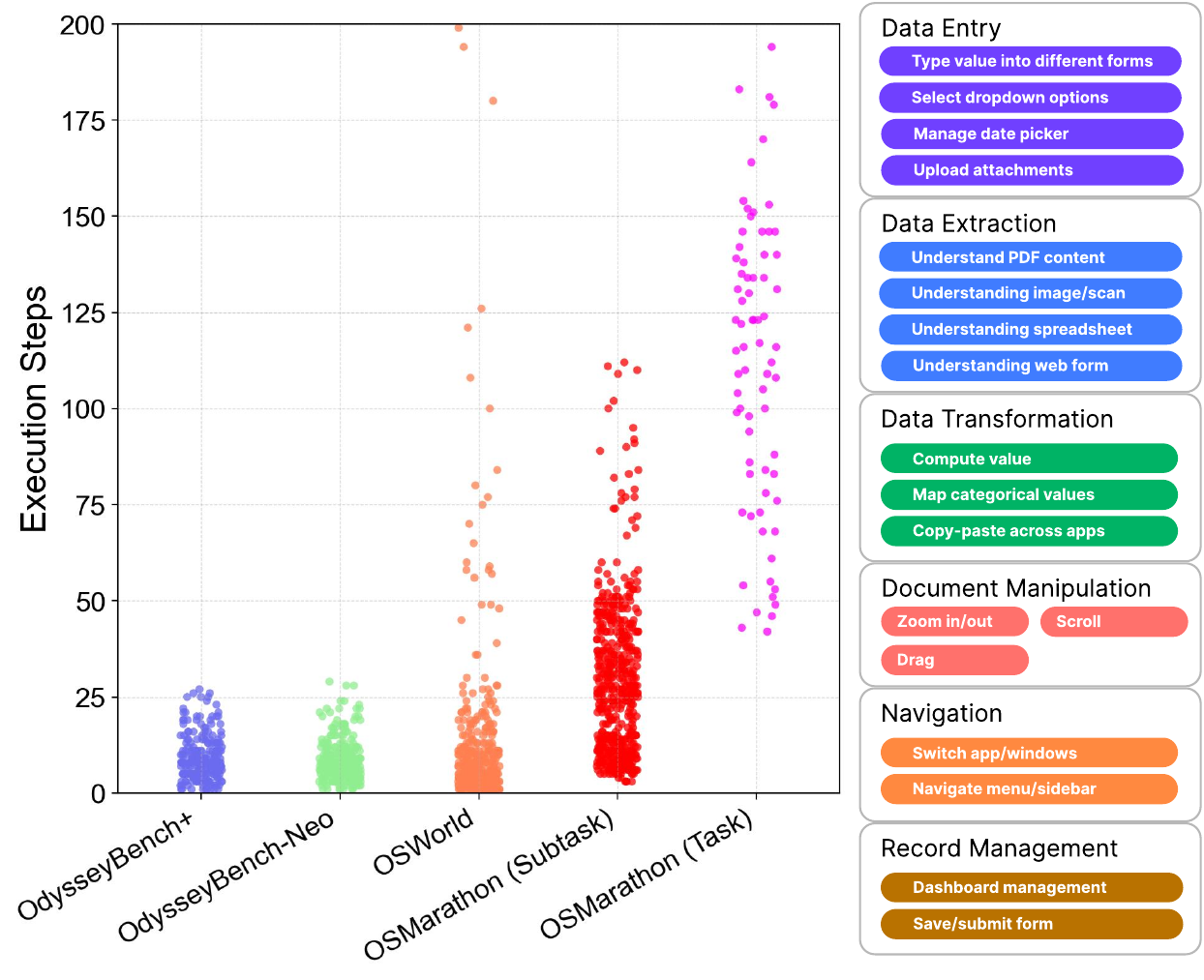}
  \caption{\textbf{Benchmark statistics.} Left: Distribution of human-execution steps required to complete tasks in OdysseyBench~\cite{Wang2025OdysseyBenchEL}, OSWorld~\cite{OSWorld}, and \ourbenchmark. Right: Atomic tasks covered by \ourbenchmark.}
  \label{fig:benchmark-stats}
  \vspace{-6mm}
\end{figure}

\subsection{Benchmark Statistics}
\label{sec:benchmark-stats}

\ourbenchmark comprises \tasknum vast-horizon repetitive tasks, spanning \scenarionum daily scenarios, Business, Academic Credentialing, Education, Note Taking, and Insurance, across \domainnum domains and \envnum execution environments.
Each task in \ourbenchmark follows the three-phase structure defined in \cref{sec:task-definition}.
All tasks in \ourbenchmark are drawn from real-world workflows and built to match this repetition feature.

\noindent\textbf{Diversity.}
\ourbenchmark exhibits diversity at the atomic-operation, subtask, and task levels. 
At the \emph{atomic-task level}, \ourbenchmark includes \atomictypenum atomic actions common in daily digital workflows, such as navigating folders or entering values into a spreadsheet.
These atomic tasks fall into \atomictaskcategories categories: data extraction, data entry, navigation, document manipulation, data transformation, and record management (\Cref{fig:benchmark-stats} (Right)). 
At the \emph{subtask level}, the data instances $d_i$ that the agent must process are themselves heterogeneous: real images and PDFs captured under varying lighting and angles, in portrait or landscape orientation, \etc.
The PDFs alone span multiple layouts (one- or two-column), document functions (flight tickets, car-rental receipts, transcripts, insurance policies), text densities (dense or loose), and capture modalities (native, scanned, printed-from-email, long screenshot).
At the \emph{task level}, individual tasks span multiple applications combined following the real-world use-case, typical tasks involve the mix of \inlinelogo{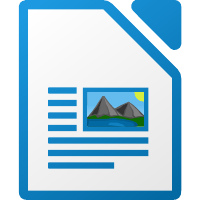} Documents, \inlinelogo{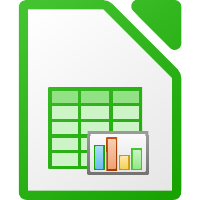} Spreadsheet, \inlinelogo{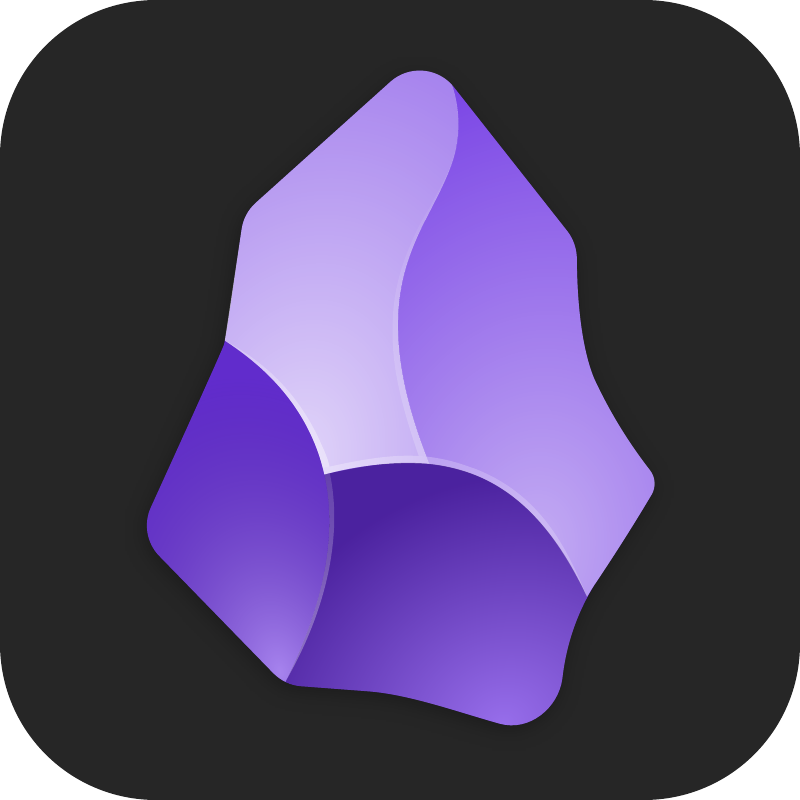} Note-taking apps, \inlinelogo{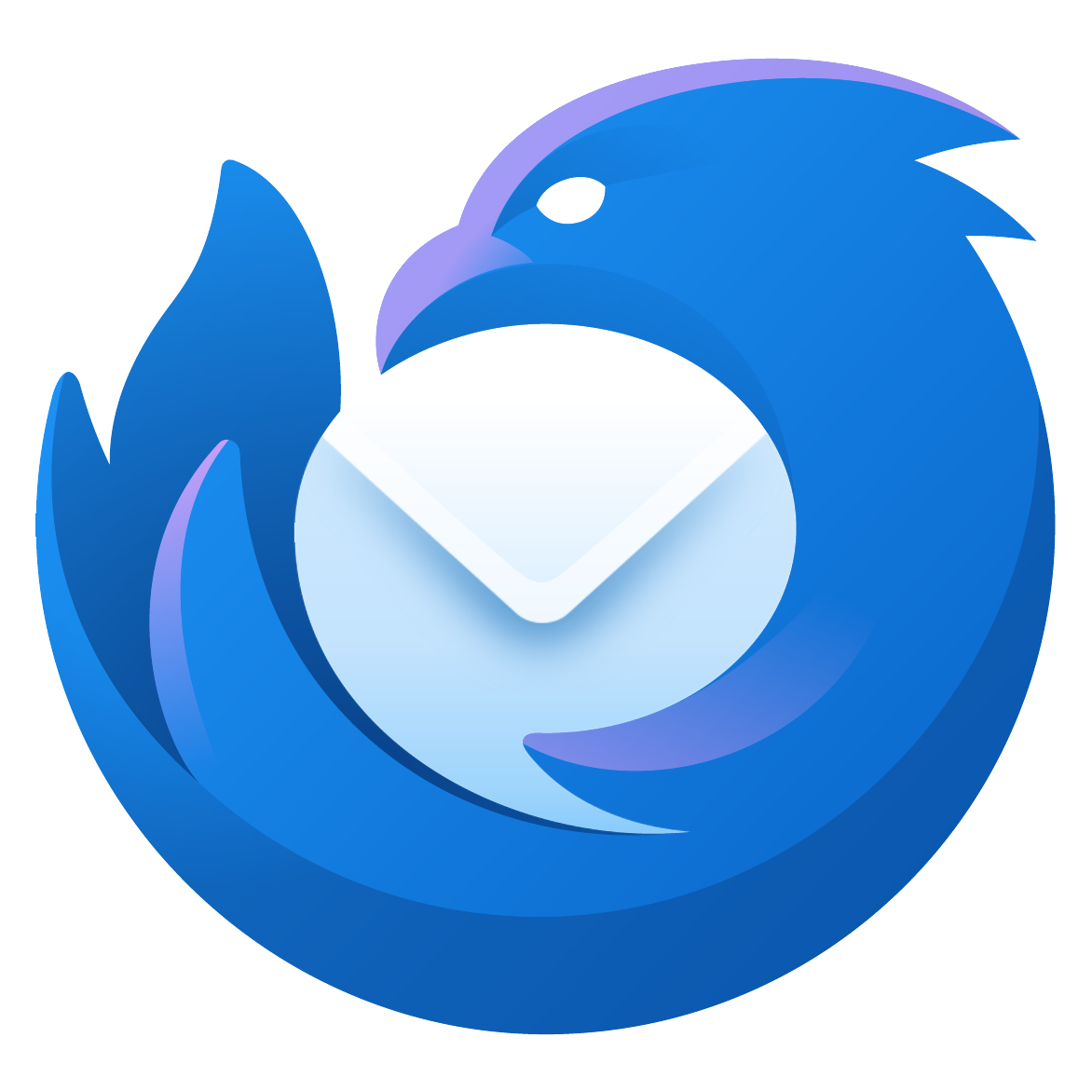} Emails, \inlinelogo{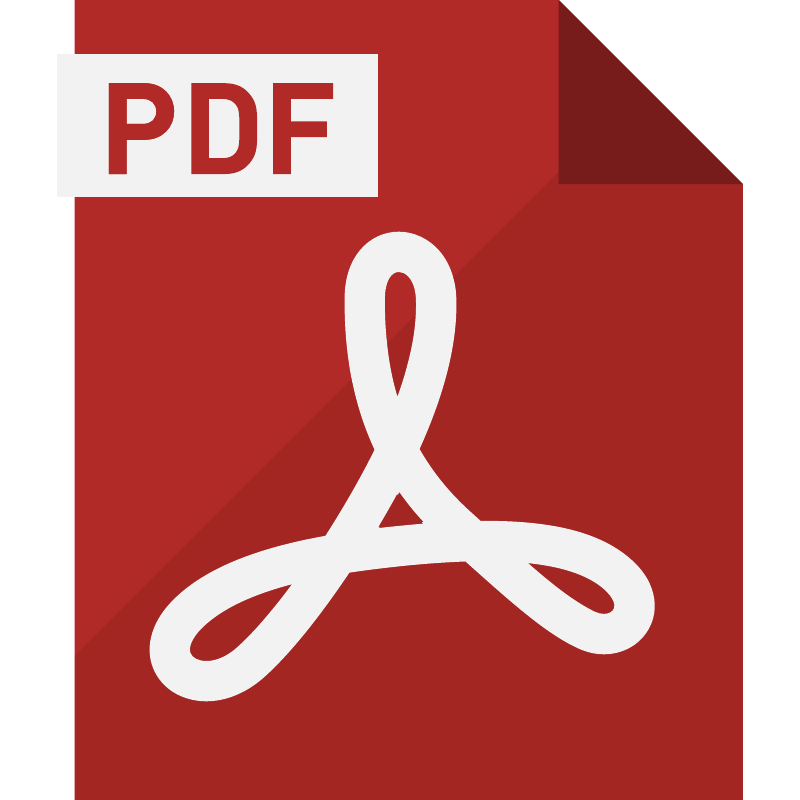} PDF, \inlinelogo{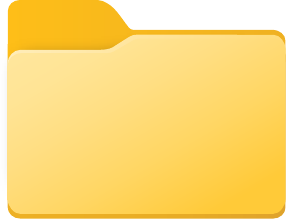} OS-level utilities, \etc.

\noindent\noindent\textbf{Subtask dependency.}
Crucially, tasks in \ourbenchmark are not simple aggregations of independent subtasks. 
Because all subtasks share a single execution environment, \textbf{the outcome of an earlier subtask directly influences the success of later ones}.
For example, for a business expense reporting case, the agent must log each receipt into the correct fields with the right amount, currency, etc. 
An earlier entry logged into the wrong field can displace a later entry, and an incorrectly completed subtask leaves the environment in a misleading state that cascades into subsequent ones. 
Furthermore, for the summary section in the form, any incorrect amount for the previous subtasks will lead to failure in the summary. 
This \textbf{error-compounding structure} is what makes \ourbenchmark a challenging benchmark for real-world, vast-horizon, repetitive tasks. 


\noindent\textbf{Comparison to other CUA benchmarks.}
We compare \ourbenchmark against two counterpart benchmarks: OSWorld~\cite{OSWorld, abhyankar2025osworldhuman}, a widely-used benchmark for general computer-use agent tasks, and OdysseyBench~\cite{Wang2025OdysseyBenchEL}, which targets long-horizon office workflows. 
\Cref{fig:benchmark-stats}~(Left) plots the required execution steps for tasks in these three benchmarks.
Subtasks in \ourbenchmark already match the horizon of full OSWorld and OdysseyBench tasks (typically $<30$ steps), with a non-trivial fraction exceeding them.
At the task level, \ourbenchmark extends well beyond either prior benchmark, spanning roughly 50 to 200 steps end-to-end, and is further scalable through the number of data instances $N$ each task processes. 
\ourbenchmark thus poses a challenge at both granularities: each subtask is itself a multi-step workflow demanding coherent reasoning and execution, and the full task adds vast-horizon consistency across all $N$ instances.






\begin{table*}[!htbp]
\tiny
\centering
\setlength{\tabcolsep}{3pt}
\renewcommand{\arraystretch}{1.1}
\resizebox{\linewidth}{!}{
\begin{tabular}{l|AAAAA|BBBB|CCC|D}
\hline
\textbf{Agents}
 & \textbf{Business}
 & \stackanchor{\textbf{Academic}}{\textbf{Cred.}}
 & \textbf{Education}
 & \stackanchor{\textbf{Note}}{\textbf{Taking}}
 & \textbf{Insurance}
 & \textbf{Web}
 & \textbf{Spreadsheet}
 & \textbf{Document} 
 & \textbf{Obsidian}
 & \textbf{2-app}
 & \textbf{3-app}
 & \textbf{4-app}
 & \textbf{Overall} \\
\hline
\textcolor{blue}{\makecell[l]{OpenCUA7B}}                       &  02.29 / 00.00 &  04.57 / 00.00 &  00.39 / 00.00 & 15.86 / 20.84 & 10.16 / 00.00 &  07.19 / 00.00 &  00.39 / 00.00 &  00.00 / 00.00 & 15.86 / 20.84 & 06.30 / 04.17 & 10.35 / 00.00 & 00.26 / 00.00 &  06.65 / 04.17 \\
\textcolor{blue}{\makecell[l]{UITARS7B}}                        &  00.70 / 00.00 &  00.00 / 00.00 &  00.00 / 00.00 & 14.47 / 19.45 &  02.34 / 00.00 &  01.50 / 00.00 &  00.00 / 00.00 &  00.00 / 00.00 & 14.47 / 19.45 & 03.68 / 04.32 & 02.34 / 00.00 & 00.00 / 00.00 &  03.50 / 03.89 \\
\textcolor{red}{\makecell[l]{AgentS2.5 w/ GPT5}}               & 47.93 / 30.61 & 40.76 / 37.08 & 02.73 / 00.00 & 40.49 / 34.71 & 35.59 / 14.82 & 50.55 / 38.88 & 05.30 / 00.41 & 30.56 / 00.00 & 40.49 / 34.71 & 50.53 / 39.89 & 26.90 / 15.44 & 11.10 / 00.00 & 33.50 / 23.44 \\
\textcolor{red}{\makecell[l]{UiPath w/ Opus 4.5}} & 47.85 / 29.95 & 34.32 / 19.08 & 15.17 / 01.56 & 34.92 / 23.93 & 27.24 / 07.29 & 40.11 / 27.46 & 24.74 / 01.04 & 23.75 / 08.68 & 34.92 / 23.93 & 41.04 / 31.47 & 37.31 / 02.95 & 18.03 / 03.94 & 31.90 / 16.36 \\
\textcolor{cyan}{\makecell[l]{GPT5.4-Mini}}                     & 19.27 / 21.41 & 02.79 / 01.00 & 02.28 / 00.00 & 15.07 / 14.37 & 10.11 / 00.00 & 12.53 / 12.80 & 05.58 / 00.74 & 08.75 / 00.00 & 15.07 / 14.37 & 13.25 / 15.67 & 11.82 / 01.11 & 04.44 / 00.00 & 09.90 / 07.35 \\
\textcolor{cyan}{\makecell[l]{GPT5.4-Nano}}                     & 00.00 / 00.00 & 00.18 / 00.00 & 00.50 / 00.00 & 14.47 / 14.29 & 03.13 / 00.00 & 01.32 / 00.00 & 00.33 / 00.00 & 00.00 / 00.00 & 14.47 / 14.29 & 02.96 / 02.86 & 03.13 / 00.00 & 00.33 / 00.00 & 03.65 / 02.86 \\
\textcolor{cyan}{\makecell[l]{o3}}                              & 00.00 / 00.00 & 00.38 / 00.50 & 00.50 / 00.00 & 19.44 / 16.50 & 03.91 / 00.00 & 01.71 / 00.20 & 00.33 / 00.00 & 00.00 / 00.00 & 19.44 / 16.50 & 04.04 / 03.50 & 03.91 / 00.00 & 00.00 / 00.00 & 04.85 / 03.40 \\
\textcolor{magenta}{\makecell[l]{GPT5.4 NCU}}  & \textbf{82.90} / \textbf{61.22} & \textbf{67.68} / \textbf{55.95} & \textbf{57.77} / \textbf{35.42} & \textbf{78.98} / \textbf{65.73} & \textbf{54.12} / \textbf{38.63} & \textbf{77.26} / \textbf{66.54} & \textbf{59.81} / \textbf{29.00} & \textbf{42.09} / \textbf{23.96} & \textbf{78.98} / \textbf{65.73} & \textbf{79.83} / \textbf{69.03} & \textbf{65.02} / \textbf{34.72} & \textbf{52.54} / \textbf{31.60} & \textbf{68.29} / \textbf{51.39} \\
\hline
\textcolor{black}{\makecell[l]{Human}}                          & \textbf{93.31} / \textbf{82.84} & \textbf{87.63} / \textbf{75.80} & \textbf{84.40} / \textbf{76.77} & \textbf{94.10} / \textbf{88.26} & \textbf{86.51} / \textbf{74.16} & \textbf{94.42} / \textbf{83.47} & \textbf{86.45} / \textbf{76.39} & \textbf{87.16} / \textbf{77.45} & \textbf{94.10} / \textbf{88.26} & \textbf{93.61} / \textbf{82.59} & \textbf{87.41} / \textbf{79.88} & \textbf{83.73} / \textbf{76.46} & \textbf{88.38} / \textbf{80.52} \\
\hline
\end{tabular}
}
\vspace{-3mm}
\caption{\textbf{Baseline performance on \ourbenchmark.} Each cell reports the field-level / subtask-level success rate (\%). Columns are grouped into four axes, distinguished by background colour: \textbf{five task scenarios} (\textcolor{chunkAtext}{light blue}), \textbf{four workflow types} (\textcolor{chunkBtext}{light yellow}), \textbf{\#apps/task} (\textcolor{chunkCtext}{light green}), and \textbf{overall} (\textcolor{chunkDtext}{light purple}). See \Cref{sec:benchmark-stats} for scenario and workflow-type definitions. Row colours mark agent categories (\cref{sec:baselines-setup}).}
\label{tab:baselines}
\vspace{-4mm}
\end{table*}

\subsection{Infrastructures \& Quality Control}
\label{sec:exec_env_data}

\paragraph{Infrastructures.}
Each task is paired with a profile of supporting data, for example, a set of receipts for a business-trip expense report, a transcript for an academic credentialing task, a batch of student exam papers for grading, a collection of PDF notes for note organisation, or a bundle of travel bookings and insurance policies for a claim. 
To produce realistic and privacy-preserving file sets for each task, we use a hybrid data pipeline.

\emph{Editable data} (digital-native PDFs such as travel bookings, insurance policies, and student papers) are generated using a \emph{format-plus-content} procedure: we collect raw documents from public online sources\footnote{\url{https://www.scribd.com/}\label{fn:scribd}}\footnote{\url{https://www.slideshare.net/}\label{fn:slideshare}} to provide realistic templates, then manually edit all task-related data (names, policy numbers, account numbers, etc.) with synthetic values to keep edits minimal to preserve the documents' real-world realism.
\emph{Uneditable data} (image receipts from existing OCR datasets~\cite{kaggleocrdata, receiptDataset} and scanned PDFs collected from the same online sources) are extended by annotating the additional fields each task requires (e.g., payment method). 
We anonymise the regions containing sensitive information to protect the privacy. 
This pipeline yields a dataset that preserves the visual and structural realism of the original documents without exposing real personal information.
The full data-collection process spanned roughly three months, and all data sourced from existing datasets is used in compliance with the original license terms.

To support rigorous evaluation of the vast-horizon, repetitive workflows defined above, we built \envnum interactive \textit{execution environments}, visualisations of which are shown in \Cref{app:appen-exeuction-example}. 
Each environment is modelled directly on a real-world daily workflow, and the set as a whole spans four workflow categories: \emph{web-based} workflows (e.g., corporate or university expense-submission portals), \emph{spreadsheet-based} workflows (e.g., grading a batch of student exam papers and recording the marks in a sheet), \emph{document-based} workflows (e.g., filling a multi-page insurance-claim form), and \emph{markdown-based} workflows (e.g., reading and organising notes in Obsidian).

\vspace{-0.25cm}
\paragraph{Quality control \& human evaluation.}
Every annotation is cross-checked by a group of three annotators to guarantee correctness. 
We then conduct a human-evaluation study with five additional evaluators, as shown in \Cref{tab:baselines} (Last Row), who were not involved in the annotation process and complete each task end-to-end under the same setup provided to the agents.
This study serves two purposes. First, as a feasibility check, it confirms that tasks in \ourbenchmark are understandable and solvable by a human, so the benchmark reflects genuine real-world difficulty rather than ambiguity or unsolvability and provides a reference for agent performance. 
Second, the human evaluation serves as a difficulty proxy: easier tasks require fewer steps and harder tasks require more. 
Based on this, we partition \ourbenchmark into two tiers, i.e. \ourbenchmarklite contains tasks humans complete within 100 steps, and \ourbenchmarkpro contains tasks requiring 100 to 200 steps.



\section{Benchmarking SOTA Agent Baselines}

\subsection{Baseline Experiment Setup}

\label{sec:baselines-setup}
\textbf{Agent baselines. }
To ensure a rigorous evaluation, we benchmark state-of-the-art VLM agents drawn from four categories: (i) \textcolor{blue}{open-weight models}, OpenCUA-7B~\cite{wang2025opencua} and UI-TARS-1.5-7B~\cite{Qin2025UI-tars}; (ii) \textcolor{red}{agentic frameworks}, AgentS2.5 with GPT-5 \cite{Agent-S2} and UIPath Screen Agent \cite{uipath2026screenagent} with Opus 4.5 \cite{anthropic2025opus45}; (iii) \textcolor{cyan}{proprietary reasoning models} from OpenAI, o3 \cite{openai2025o3}, GPT-5.4-mini~\cite{openai2026gpt54mininano}, and GPT-5.4-nano \cite{openai2026gpt54mininano}; and (iv) an \textcolor{magenta}{agent with native computer use}, GPT-5.4 Native Computer Use \cite{openai2026gpt54}, with high reasoning effort.

\noindent\textbf{Evaluation protocol.}
\label{sec:metrics}
We evaluate agents under budgets matched to human performance, following the evaluation protocol of OSWorld~\cite{OSWorld}. 
To enable fine-grained analysis of the results, we report success rates at different granularity levels: (i) \emph{field-level}, the correctness of each individual element (e.g., a single form field); (ii) \emph{subtask-level}, the success rate over the group of individual elements that constitute one subtask (e.g., one receipt entered correctly in full). 
Due to the space limit, we provide further evaluation of the baseline performance on the end-to-end \emph{task-level} success rate in \Cref{sec:task-sr}.


\subsection{Results \& Analysis}
\label{sec:baseline-analysis}
\Cref{tab:baselines} reports field- and subtask-level success rates on \ourbenchmark, broken down by scenario, workflow type, and number of apps per task.

\begin{figure}[!htbp]
\centering
\vspace{-0.2cm}
\begin{takeawaybox}[width=\columnwidth]
\textbf{Obs. 1}: Humans still largely outperform SOTA agents. \\
\textbf{Obs. 2}: Best product-level native agent still struggles to solve vast-horizon, repetitive tasks. 
\end{takeawaybox}
\vspace{-0.6cm}
\end{figure}



\begin{figure*}
    \centering
    \includegraphics[width=\linewidth]{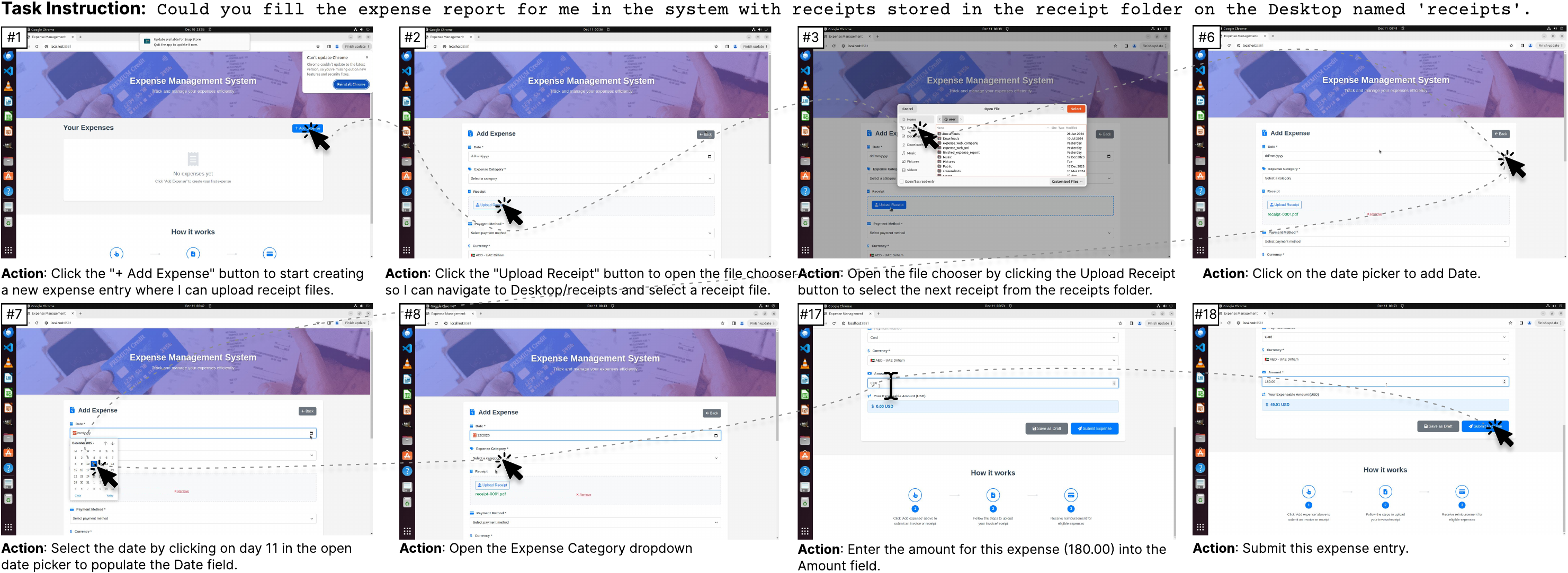}
    \vspace{-5mm}
    \caption{\textbf{Failure case analysis: AgentS2.5 + GPT-5.} 
    The agent clicks the salient \emph{Upload Receipt} button (\#2) before reading any receipt, drawn by the LLM training association between \emph{Upload Receipt} and \emph{Fill an Expense Report}.
    Once on this trajectory, the agent hallucinates plausible values for the date (\#7) and payment amount (\#17) with no grounded source to draw from, compounding one planning error into a cascade of incorrect entries.
    }
    \label{fig:failure-case}
    \vspace{-5mm}
\end{figure*}

\noindent\textbf{Bottleneck.} 
OpenCUA-7B~\cite{wang2025opencua}, UI-TARS-7B~\cite{Qin2025UI-tars}, o3~\cite{openai2025o3}, and GPT-5.4-nano~\cite{openai2026gpt54mininano} are bottlenecked at the atomic-operation level.
The cause is not solely in raw grounding ability, but also in \emph{high-level user-intention understanding}: the agent is unable to reliably decompose a high-level intention such as \emph{file an expense report} into the concrete atomic steps it implies (e.g., opening a receipt, reading the payment amount, and entering it into the form).
The remaining agents are able to execute atomic actions but bottleneck at the subtask and task levels, where we identify two failure modes.
(i) \textbf{\emph{Visual-salience bias and downstream hallucination.}} 
As shown in \Cref{fig:failure-case} (Step \#2), the agent's reasoning tends to be affected by the UI element that is most visually salient and most directly named in the instruction. 
This is likely related to how the model was pre-trained.
\emph{Expense report} and \emph{upload receipt} commonly appear together in LLM training data, so during reasoning, the agent easily associates the two and jumps to \emph{upload receipt} directly. 
The actions a global plan would require, e.g., first opening a receipt to read its content, are less directly linked to \emph{expense report} in training, so the agent is more easily distracted by \emph{upload receipt} and less likely to perform them.
(ii) \textbf{\emph{Weak global planning and persistence.}} 
Even when each subtask is executed correctly in isolation, agents fail to sustain the full $N$-instance loop.
Without a global plan, they process data instances opportunistically rather than in a planned order, and the loop often terminates before all $N$ instances are completed.
We elaborate the analysis in \Cref{app:analysis-levels}. 



\vspace{-2mm}
\noindent\textbf{Analysis by scenario and workflow type. }
Across \textbf{scenarios}, education and insurance emerge as uniformly the hardest, which we attribute to their stronger \emph{cross-referencing} dependencies: each subtask refers to information from multiple sources (e.g., an insurance claim must reference both the policy and the supporting documents), which is less intuitive for agents, making hallucination more likely.
\Cref{tab:baselines} (\textcolor{chunkCtext}{light green}) also confirms this, when tasks are grouped by \#apps each must cross-reference, both field-level and subtask-level success drop sharply and near-monotonically as the \#app/task grows.
Across \textbf{workflow} types, performance varies with \emph{error tolerance of the environment}: web workflows are more manageable because their actions are bounded and reversible, a mistake rarely renders a destructive result, while spreadsheet and document workflows expose a wider editing surface where a single destructive action (a stray drag, a misdirected paste) can corrupt the environment beyond recovery.
More analysis is provided in \Cref{app:analysis-workflow}. 

\noindent\textbf{Cost and efficiency analysis.}
\Cref{fig:cost_step_accu_method} (In Appendix) plots agent token usage, monetary cost, and step count against subtask-level accuracy, from which three observations stand out.
\textit{First}, the cheapest agent that achieves non-trivial subtask-level performance costs roughly \$5\textasciitilde\$10 per task on average (e.g. AgentS2.5 + GPT5 in \Cref{fig:cost_step_accu_method} (Middle)), and this figure scales linearly with the number of data instances $N$, already prohibitive for everyday personal use.
Two open questions follow from this. 
How can competitive performance be delivered from cheaper open-weight models? 
And how can a single human demonstration be used to personalise off-the-shelf models in a way that meaningfully reduces inference cost rather than merely improving accuracy at the same cost?
\textit{Second}, cost and accuracy are not tightly correlated across agents: UiPath~+~Opus~4.5 incurs a cost comparable to GPT-5.4 (high) but is substantially outperformed by it across all scenarios.
\textit{Third}, stronger agents are also more step-efficient: GPT-5.4 with native computer use achieves the highest success rate while taking fewer steps than several substantially weaker baselines.

\section{Agent Personalisation with \ourmethod}

The baselines evaluated above are general-purpose computer-use agents, yet real-world, vast-horizon, repetitive workflows are highly personalised: each organisation has its own expense form, teachers in different countries follow different grading standards and conventions, and each insurer maintains its own claim portal and forms, \etc. 
Adapting a general-purpose agent to a personalised workflow typically requires collecting supervised data and fine-tuning the underlying model, both prohibitively resource-intensive for individual users and small teams.

These costs, however, are largely avoidable for the workflows we target. 
Vast-horizon, repetitive tasks in \ourbenchmark feature a repetitive structure, which opens a cost-friendly alternative: subtasks within a task share a similar execution logic, so a single human demonstration of a few subtasks can carry enough information for the agent to handle the rest.
This raises the question we explore next: \emph{how to use a single human demonstration to adapt an off-the-shelf agent to a personalised, vast-horizon, repetitive workflow?}

A straightforward approach is to inject the full demonstration trajectory, e.g., every screenshot and every action, into the agent's context. 
However, this approach fails for two reasons in this case. 
(i)~\emph{\textbf{Context overflow.}} 
Demonstrations for vast-horizon, repetitive tasks far exceed agents' context window: even a demonstration covering only a single subtask spans roughly 30 steps and 9M multimodal tokens.
(ii)~\emph{\textbf{Granularity mismatch.}} Feeding the entire trajectory dilutes the useful signal: the agent only needs the slice relevant to its current step.

\begin{figure}
    \centering
    \includegraphics[width=\linewidth]{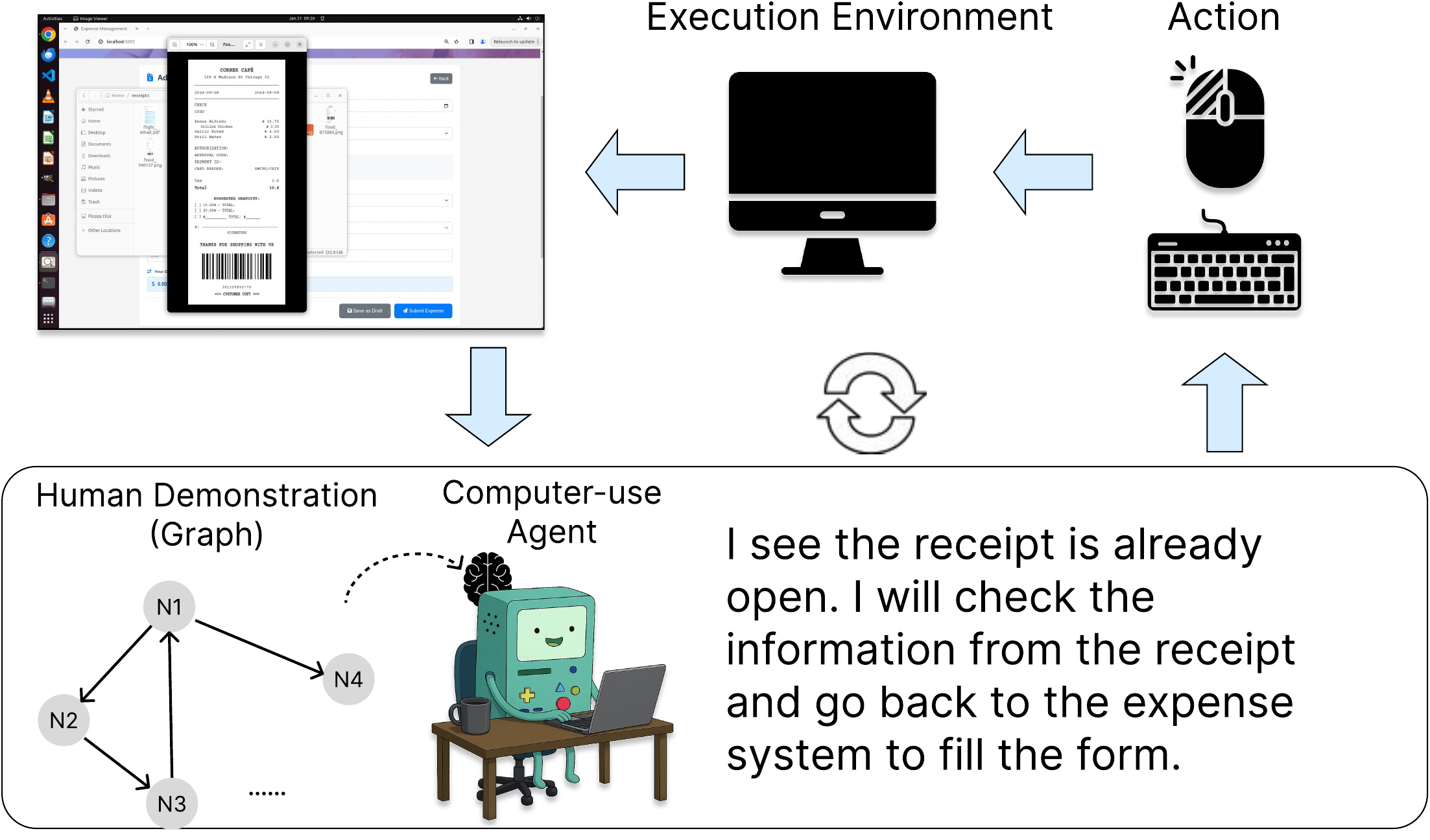}
    \caption{\textbf{\ourmethod pipeline}. A human demonstration is represented as a causal graph and incorporated into the agent’s context to guide agents' reasoning.}
    \label{fig:method}
    \vspace{-6mm}
\end{figure}

\subsection{\ourmethod}

\noindent\textbf{Graph-based demonstration with on-demand retrieval.}
To resolve both issues, we represent the demonstration as a causal graph, termed as \ourmethod (\Cref{fig:method}). 
Each node in the graph represents one demonstrated step and stores two density levels, a textual description for compact retrieval and the screenshots for detailed visual context, while directed edges carry the executed action information. 
The textual abstraction of the graph is placed in the agent's system prompt as a persistent workflow overview, giving the agent direct visibility over all candidate nodes.
At each step, the agent retrieves the demonstration node best matching its planned action, along with its neighbours; the multimodal content of these nodes is injected into context during the reasoning of the next step.
The full demonstration thus remains usable within the agent's context window, while at any step the agent sees only the slice of the demonstration relevant to its action, addressing both the context-overflow and granularity-mismatch problems of the full-trajectory approach.
We cover more details in \Cref{app:method-details}.

\subsection{Experiments \& Analysis}

\begin{table}[!htbp]
\vspace{-2mm}
\centering
\setlength{\tabcolsep}{3pt}
\renewcommand{\arraystretch}{1.1}
\resizebox{\linewidth}{!}{
    \begin{tabular}{l|ccccc|c}
    \toprule
    \bfseries Agents & \bfseries Business & \bfseries \makecell{Academic\\Cred.} & \bfseries Education & \bfseries \makecell{Note\\Taking} & \bfseries Insurance & \textbf{Overall} \\
    \midrule
    S2.5 + GPT5                               & 51.29 / 39.97 & 40.76 / 37.08 &  1.37 /  0.00 & 40.49 / 34.71 & 35.59 / 14.82 & 33.90 / 25.31 \\
    S2.5 + GPT5 + Orch.       & 36.87 / 19.41 & 26.74 / 19.99 &  0.37 /  0.00 & 32.75 / 25.96 & 48.61 / 14.71 & 29.07 / 16.01 \\
    \hline
    \textbf{S2.5 + GPT5 + \ourmethod}     & \textbf{81.10 / 60.36} & \textbf{48.19 / 46.65} & \textbf{20.97 / 16.17} & \textbf{53.23 / 46.67} & \textbf{51.25 / 36.36} & \textbf{50.95 / 41.24} \\
    \makecell[l]{S2.5 + GPT5\\ \hspace{1em} + Orch. + \ourmethod} & 64.12 / 42.80 & 43.26 / 41.33 & 13.45 /  0.00 & 38.55 / 38.48 & 44.53 / 14.38 & 40.78 / 27.40 \\
    \bottomrule
    \end{tabular}
}
\vspace{-2mm}
\caption{\textbf{Orchestrator-based task decomposition alone is counterproductive; \ourmethod improves performance in both settings.} Variants on AgentS2.5 + GPT-5: baseline, with orchestrator (+\,Orch.), with \ourmethod alone, and with orchestrator + \ourmethod. Each cell reports field / subtask-level success rate (\%).}
\label{tab:orchestrator}
\vspace{-6mm}
\end{table}

\noindent\textbf{Agent personalisation experiment setup.}
We conduct the agent-personalisation experiments on AgentS2.5 with GPT-5 \cite{Agent-S2}, which demonstrates solid atomic-level competence in our baseline results and therefore offers the headroom required to study higher-level personalisation effects. 
To investigate whether the difficulty of \ourbenchmark can be reduced by explicit task decomposition, we adopt a complementary baseline: wrapping the agent as a solver in a task orchestrator that breaks the workflow into per-instance subtasks and assigns them to a solver agent one at a time. 
We then apply the graph-based demonstration to both setups, the vanilla agent and the orchestrator-wrapped agent, and evaluate all variants across \ourbenchmark under the same protocol as the baseline experiments. 
We report the quantitative results in \Cref{tab:orchestrator} and cover the additional results, analysis and implementation details for the orchestrator and demonstration integration are provided in \Cref{app:persoanlisation-imple-details}.


\begin{figure}[!htbp]
\centering
\vspace{-0.2cm}
\begin{takeawaybox}[width=\columnwidth]
\textbf{Obs. 3}: Naively applying the orchestrator framework is largely counterproductive, providing only occasional gains, e.g., on the field-level metric for the insurance scenario.
\end{takeawaybox}
\vspace{-0.6cm}
\end{figure}
\begin{figure*}[!t]
    \centering
    \includegraphics[width=.96\linewidth]{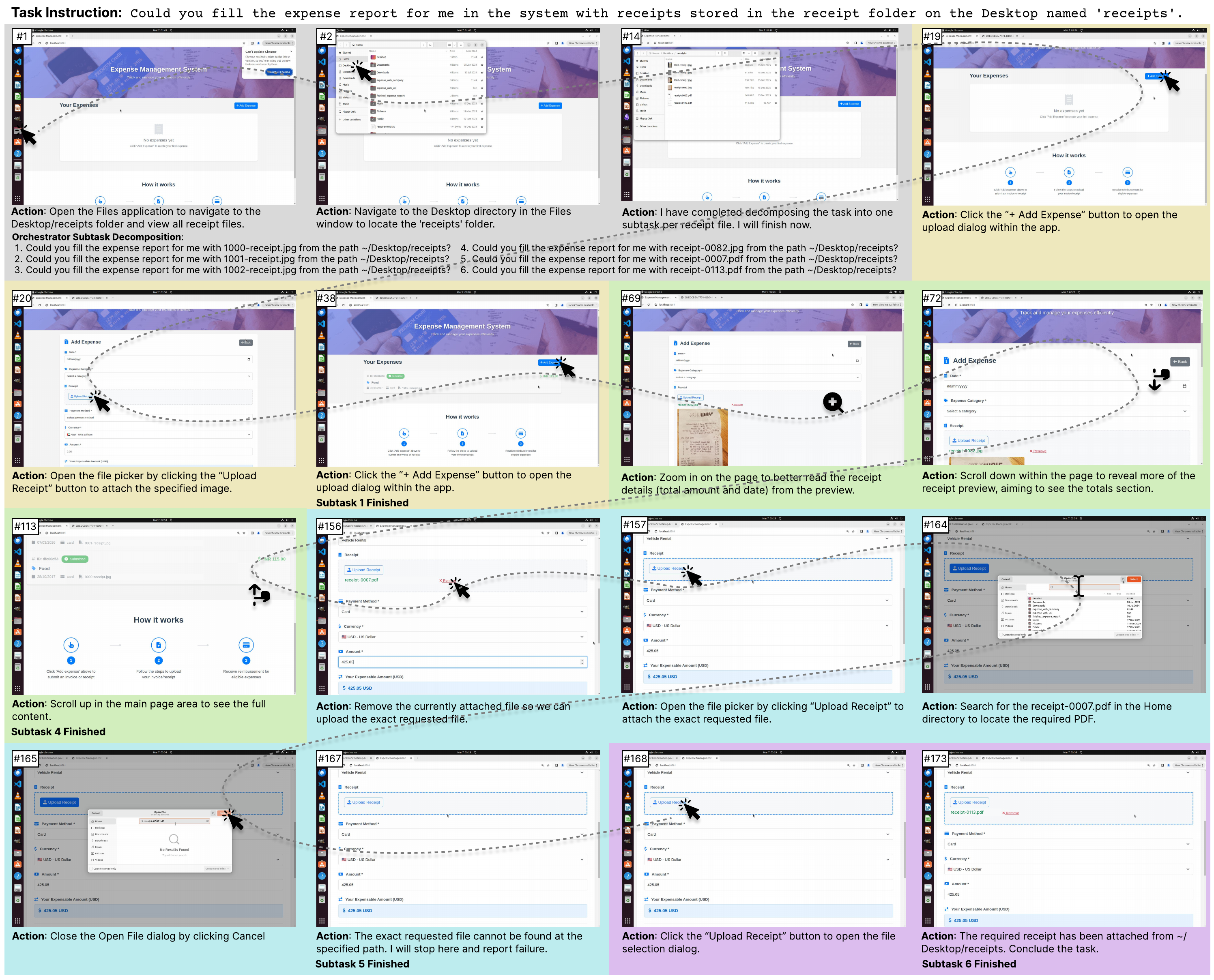}
    \caption{
    \textbf{Failure case analysis: AgentS2.5 + GPT-5 + Orchestrator.} The framework completes the first four subtasks successfully but drifts on the fifth: the solver mistakenly removes the uploaded receipts (\#156) and fails to recover, leaving the environment in a corrupted state (\#167). The next solver, initialised on this state, misreads its instruction as a request to simply upload the receipt (\#168) and finishes the task incorrectly. Screenshot backgrounds denote the role of each agent: \textcolor[RGB]{66,66,66}{grey} for the task orchestrator $\mathbf{O}$, \textcolor{chunkBtext}{light yellow} for solver $\mathbf{S}_1$, \textcolor[RGB]{93,255,13}{light green} for $\mathbf{S}_4$, \textcolor{cyan}{cyan} for $\mathbf{S}_5$, and \textcolor{Violet}{light purple} for $\mathbf{S}_6$. $\mathbf{S}_2$ and $\mathbf{S}_3$ are omitted as they execute normally and are not part of the failure cascade.
    }
    \vspace{-6mm}
    \label{app:orchestrator-failure-case}
\end{figure*}
\noindent\textbf{Orchestrator decomposition analysis. }
The orchestrator mitigates task-level persistence, but the subtask-internal logic, which our analysis in \cref{sec:baseline-analysis,app:analysis-levels} identifies as a severe bottleneck, falls outside its scope.
Furthermore, the framework introduces a new failure mode: a solver agent's mistake corrupts the shared environment, and the next solver agent, with no context of the error, reasons from this corrupted state and compounds the failure.


We provide a failure case analysis of this in \Cref{app:orchestrator-failure-case}. 
The framework completes the first four subtasks successfully (Step \#113) but drifts on the fifth: due to the incorrect subtask logic, the solver agent decides to remove the uploaded receipts (Step~\#156), believing them incorrect. 
When attempting to re-upload, it searches the wrong folder (\texttt{\textbf{home}}) (Step~\#164), fails to find the file (Step~\#165), and reports \texttt{\textbf{FAIL}}, leaving the environment in a corrupted state (Step~\#167). 
The next solver agent, initialised on this corrupted state with all numerical fields already filled, sees \texttt{Upload Receipt} as the only remaining action and misreads its instruction, \emph{``Could you fill the expense report for me with receipt-0113.pdf from \textasciitilde/Desktop/receipts?''}, as a request to simply upload the receipt. 
It re-uploads (Step~\#168) and reports subtask 6 as completed, finishing the task incorrectly.

\begin{figure}[!htbp]
\centering
\vspace{-0.4cm}
\begin{takeawaybox}[width=\columnwidth]
\textbf{Obs. 4}: Graph human demonstration improves general CUAs on vast-horizon, repetitive tasks, lifting weaker agents close to SOTA performance at a low cost.
\end{takeawaybox}
\vspace{-0.6cm}
\end{figure}

\noindent\textbf{\ourmethod analysis.} 
\ourmethod improves both field-level and subtask-level success rates on the vanilla agent by guiding it through the correct execution logic. 
This lifts a weaker agent (AgentS2.5 + GPT-5) close to the performance of an SOTA native agent (GPT-5.4 with native computer use), giving general-purpose agents a substantial performance boost.
When applied to the orchestrator framework, \ourmethod is still helpful: Orchestrator + Demo outperforms the Orchestrator alone, but its effect is weaker than in the vanilla setup, and in some cases falls below the vanilla agent itself. 
This is because \ourmethod acts as a soft guide, and the agent can still deviate. 
In this setting, when a solver agent's mistake corrupts the shared environment, even with the demonstration guide, the next solver agent still lacks context of the error and reasons from this corrupted state, compounding the failure.


\section{Conclusion}
\vspace{-0.2cm}
This paper investigates vast-horizon, repetitive CUA tasks: workflows that process data instances iteratively over vast horizons.
We introduce \ourbenchmark, the first benchmark for this task type, spanning \tasknum tasks across \scenarionum scenarios and \domainnum domains, and establish baselines on leading current agents.
To bridge the observed performance gaps, we explore a cost-friendly personalisation strategy, \ourmethod, that encodes a single human demonstration as a causal graph and retrieves only step-relevant information into context.
Extensive experiments demonstrate both the real-world challenges in this task and the effectiveness of \ourmethod, laying a foundation for future CUA research on vast-horizon, repetitive tasks.

\section*{Limitations}
The human demonstration acts as a soft guide for personalisation: nothing constrains the agent to follow it consistently, and when the agent deviates, the demonstration cannot help it recover. Personalising general CUAs for vast-horizon, repetitive workflows, therefore, remains an open problem worth exploring.

\section*{Acknowledgments}
This work was conducted during an internship at Microsoft; we thank Microsoft Research (MSR) and Windows Cloud Experience (WCX) for their support. We also thank the authors of the OSWorld benchmark for their open-source infrastructure, which served as a foundation for this project.






\bibliography{custom}

\clearpage

\appendix

\section{Appendix}
\label{app:appendix}
\subsection{More Analysis}
\label{app:more-results}

\subsubsection{Task-level Success Rate Result}

\label[appendix]{sec:task-sr}
We run our experiments on a local VMWare virtual machine. 
For the open weights models, we host them using the vLLM \cite{kwon2023vllm} framework.

\begin{table}[!htbp]
\small
\centering
\setlength{\tabcolsep}{8pt}
\begin{tabular}{lc}
\toprule
\bfseries Agent & \bfseries Task-level SR \\
\midrule
\textcolor{blue}{\makecell[l]{OpenCUA7B}}    & \phantom{0}0.00\% \\
\textcolor{blue}{\makecell[l]{UITARS7B}}   & \phantom{0}0.00\% \\
\textcolor{red}{\makecell[l]{AgentS2.5 w/ GPT5}}   & \phantom{0}2.78\% \\
\textcolor{red}{\makecell[l]{UiPath w/ Opus 4.5}} & \phantom{0}0.56\% \\
\textcolor{cyan}{\makecell[l]{GPT5.4-Mini}}    & \phantom{0}0.00\% \\
\textcolor{cyan}{\makecell[l]{GPT5.4-Nano}}    & \phantom{0}0.00\% \\
\textcolor{cyan}{\makecell[l]{o3}}   & \phantom{0}0.00\% \\
\textcolor{magenta}{\makecell[l]{GPT5.4 Native Computer Use}}   & 11.53\% \\
\midrule
\textbf{Human}                  & \textbf{82.62\%} \\
\bottomrule
\end{tabular}
\vspace{2mm}
\caption{End-to-end task success rate for baseline agents. Agent colours follow the four categories in \cref{sec:baselines-setup}.}
\label{app:task-sr}
\end{table}

\Cref{app:task-sr} reports the end-to-end task-level success rate of all baselines. Finishing a vast-horizon, repetitive task end-to-end is highly challenging for current agents: only the strongest native computer-use agent (GPT-5.4 Native Computer Use) achieves a non-trivial success rate ($11.53\%$), while every other baseline falls below $3\%$. 
Human evaluators, in contrast, complete $82.62\%$ of tasks, indicating that the gap reflects agent capability limits rather than task infeasibility.
The low task-level success rate confirms that current agents struggle on all three capabilities required for a vast-horizon, repetitive task (\Cref{fig:teaser}, right): high-level planning, consistent iterative execution, and the implicit closing actions. Failure at any one collapses end-to-end success.

\begin{figure*}[!htbp]
    \centering
  \includegraphics[width=\linewidth]{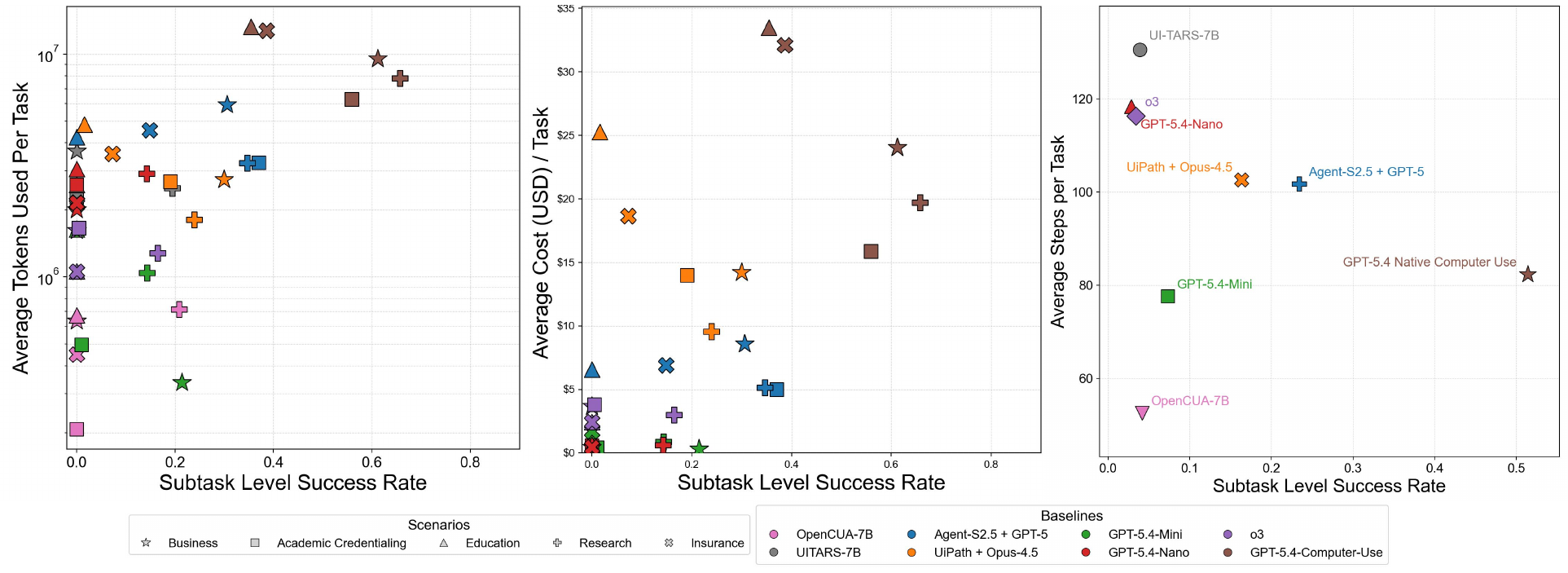}
  \caption{\textbf{Cost, efficiency and performance relationship across baseline agents on \ourbenchmark.} Subtask-level success rate plotted against token usage (left), monetary cost in USD \$ (middle), and step count (right).}
  \label{fig:cost_step_accu_method}
\end{figure*}


\subsubsection{Analysis Across Difficulty Tiers}


\Cref{app:results-difficulty-split} reports baseline performance on the two tiers. 
A clear difficulty gap is visible across all three metrics: every baseline achieves higher field-level, subtask-level, and task-level SR on \ourbenchmarklite than on \ourbenchmarkpro, confirming that the difficulty partition produces a meaningful performance distinction. 
The strongest agents can occasionally complete \ourbenchmarklite tasks end-to-end, i.e. AgentS2.5, UiPath, and GPT5.4 NCU reach task-level SR of $5.56\%$, $1.11\%$, and $16.22\%$, but drop sharply on \ourbenchmarkpro ($0\%$, $0\%$, and $6.83\%$). 
Weaker agents fail to complete tasks end-to-end on either tier, yet the difficulty gap remains visible at the field and subtask levels (e.g., OpenCUA7B: $10.18\% \to 3.13\%$ field-level).

\begin{table*}[t]
\scriptsize
\centering
\resizebox{\linewidth}{!}{
\begin{tabular}{l|ccc|ccc}
\toprule
\multirow{2}{*}{\bfseries Agents}    & \multicolumn{3}{c|}{\bfseries \ourbenchmarklite} & \multicolumn{3}{c}{\bfseries \ourbenchmarkpro} \\
\cmidrule(lr){2-4} \cmidrule(lr){5-7}
            & Field-level SR  & Subtask-level SR  &  Task-level SR & Field-level SR  & Subtask-level SR  & Task-level SR \\
\midrule
\textcolor{blue}{\makecell[l]{OpenCUA7B}}   & 10.18\%    & 07.00\%     & 00.00\%     & 03.13\%     & 01.33\%     & 00.00\%    \\
\textcolor{blue}{\makecell[l]{UITARS7B}}     & 06.24\%     & 06.44\%     & 00.00\%     & 00.77\%     & 01.33\%     & 00.00\%    \\
\textcolor{red}{\makecell[l]{AgentS2.5 w/ GPT5}}    & 36.07\%    & 26.33\%    & 05.56\%     & 30.38\%    & 20.55\%    & 00.00\%    \\
\textcolor{red}{\makecell[l]{UiPath w/ Opus 4.5}}  & 34.06\%    & 20.58\%    & 01.11\%     & 29.74\%    & 12.15\%    & 00.00\%    \\
\textcolor{cyan}{\makecell[l]{GPT5.4-Mini}}        & 11.10\%    & 08.33\%     & 00.00\%     & 08.70\%     & 06.38\%     & 00.00\%    \\
\textcolor{cyan}{\makecell[l]{GPT5.4-Nano}}     & 04.16\%     & 03.12\%     & 00.00\%     & 03.15\%     & 02.59\%     & 00.00\%    \\
\textcolor{cyan}{\makecell[l]{o3}}   & 05.45\%     & 03.68\%     & 00.00\%    & 04.24\%     & 03.12\%     & 00.00\%        \\
\textcolor{magenta}{\makecell[l]{GPT5.4 NCU}}        & 70.12\%    & 54.85\%    & 16.22\%    & 66.46\%    & 47.93\%    & 06.83\%   \\
\bottomrule
\end{tabular}
}
\caption{\textbf{Baseline performance comparison between the two \ourbenchmark difficulty tiers.} We report the field-level, subtask-level, and task-level success rates (SR) on \ourbenchmarklite and \ourbenchmarkpro. Agent colours follow the four categories in \cref{sec:baselines-setup}.}
\label{app:results-difficulty-split}
\end{table*}


\subsubsection{Analysis on Subtask/Task Levels}
\begin{figure*}[!t]
    \centering
    \includegraphics[width=\linewidth]{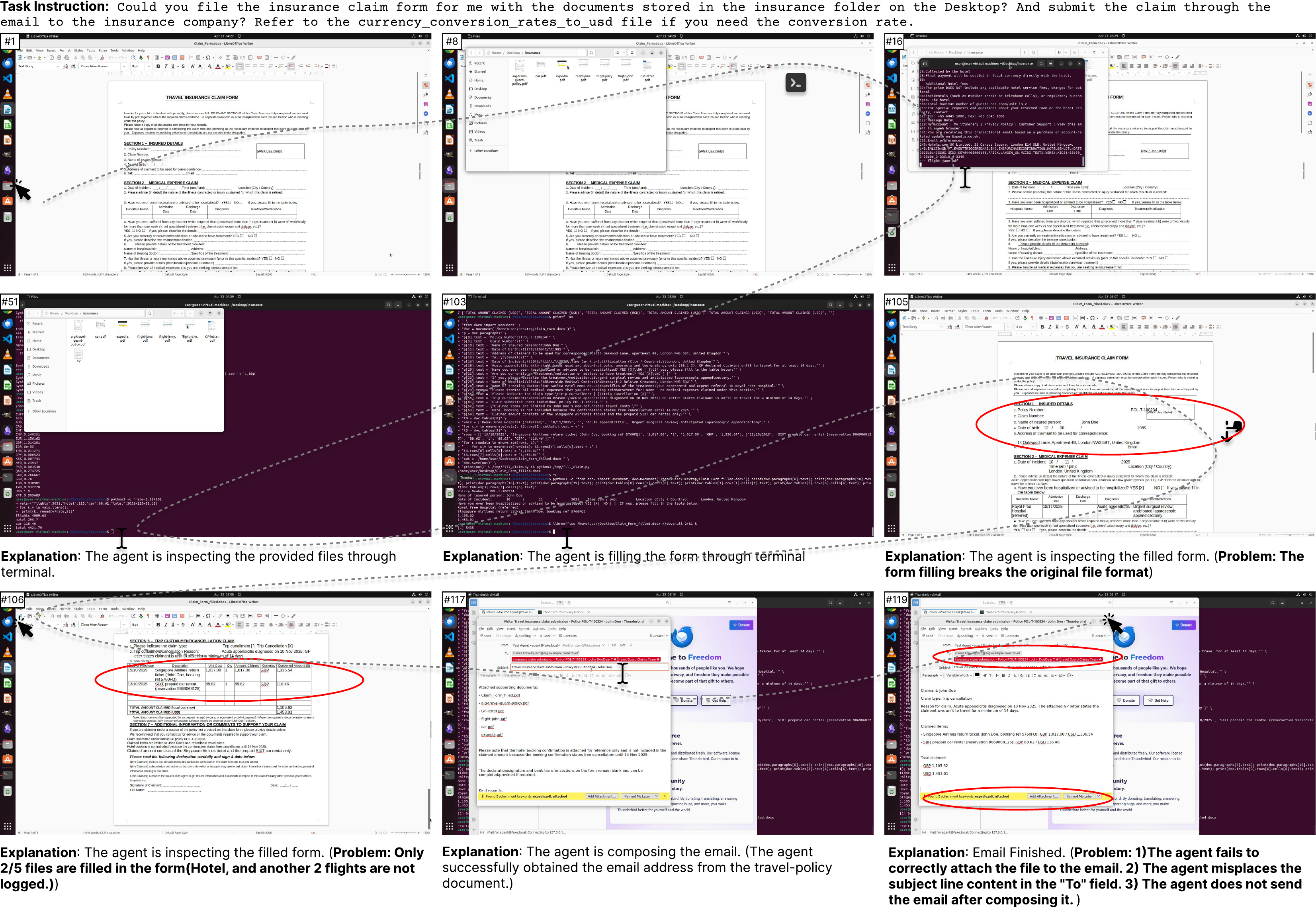}
    \caption{
    \textbf{Failure case analysis: GPT5.4 Native Computer Use (high reasoning).} 
    The agent only processes two out of five files (subtasks) and terminates prematurely (\#106).
    In the post-subtask phase, GPT5.4 attempts the closing actions but fails on three implicit requirements: it does not attach the relevant files, which is implied implicitly, misplaces the subject content in the \emph{To} field, and does not send the email after composing (\#119).
    }
    \label{fig:failure-case-gpt54}
\end{figure*}

\label[appendix]{app:analysis-levels}
We elaborate on \textbf{weak global planning and persistence}, identifying three common issues across baselines.
First, \textbf{agents lack an ordered processing strategy}: rather than committing to a deterministic order (e.g., file-listing order or chronological by receipt date), they pick the next instance opportunistically, making it hard to verify exhaustive coverage.
Second, \textbf{agents terminate prematurely}, declaring the task complete before all $N$ instances are processed. 
As shown in \Cref{fig:failure-case-gpt54}, GPT5.4 Native Computer Use is required to fill the insurance form with the supporting files provided. 
There are five supporting files to fill out, but the agent only managed to add two out of five to the form, and it proceeded to email the form. 
This failure often compounds with visual-salience bias.
For example, in an expense report case, after the agent submits a single invoice, the system displays a confirmation message such as \emph{"Invoice submitted successfully"}, which the agent may misread as confirmation that the entire workflow is complete.
Third, \textbf{implicit post-subtask actions that fall outside any single subtask, such as filling out the category-wise expense summary, are frequently skipped, especially when the environment imposes them \emph{implicitly}}: the agent must discover them through exploration, recognise them as task requirements, and incorporate them into its plan.
Together, these issues compound across the vast horizon, exposing why such tasks are challenging: completing them coherently requires global planning and persistence that current agents lack.

\subsubsection{Analysis on Different Workflow Types}
\label[appendix]{app:analysis-workflow}

The \textcolor{chunkBtext}{light yellow} columns of \Cref{tab:baselines} group the results by workflow type, revealing a clear two-tier pattern: web and obsidian workflows are more manageable for most agents, whereas spreadsheet and document workflows pose a substantially larger challenge. 
Besides the analysis in \Cref{sec:baseline-analysis}, we further analyse the observation of agents' performance on obsidian tasks. 
The obsidian workflow asks the agent to organise paper-reading notes following a provided example: creating markdown notes in the appropriate folders, moving PDFs into matching directories, and structuring annotations in the example's format.
Performance here varies with the granularity of the operations the agent must perform.
Agents reliably handle the explicit, OS-level structure, i.e. moving PDFs into the correct folders, creating notes under the matching directory layout. 
These operations are within reach even for weaker models such as OpenCUA-7B and o3, which is why those models score higher on this workflow than on others.
However, they struggle with the finer content-level conventions the example demonstrates, such as the formatting of annotations within markdown notes and the backlinks between related notes.

\subsection{Agent Personalisation Details}
\label[appendix]{app:persoanlisation-imple-details}

\subsubsection{Orchestrator Framework}
\label[appendix]{app:orchestrator-details}

\begin{figure*}[!t]
    \centering
    \includegraphics[width=\linewidth]{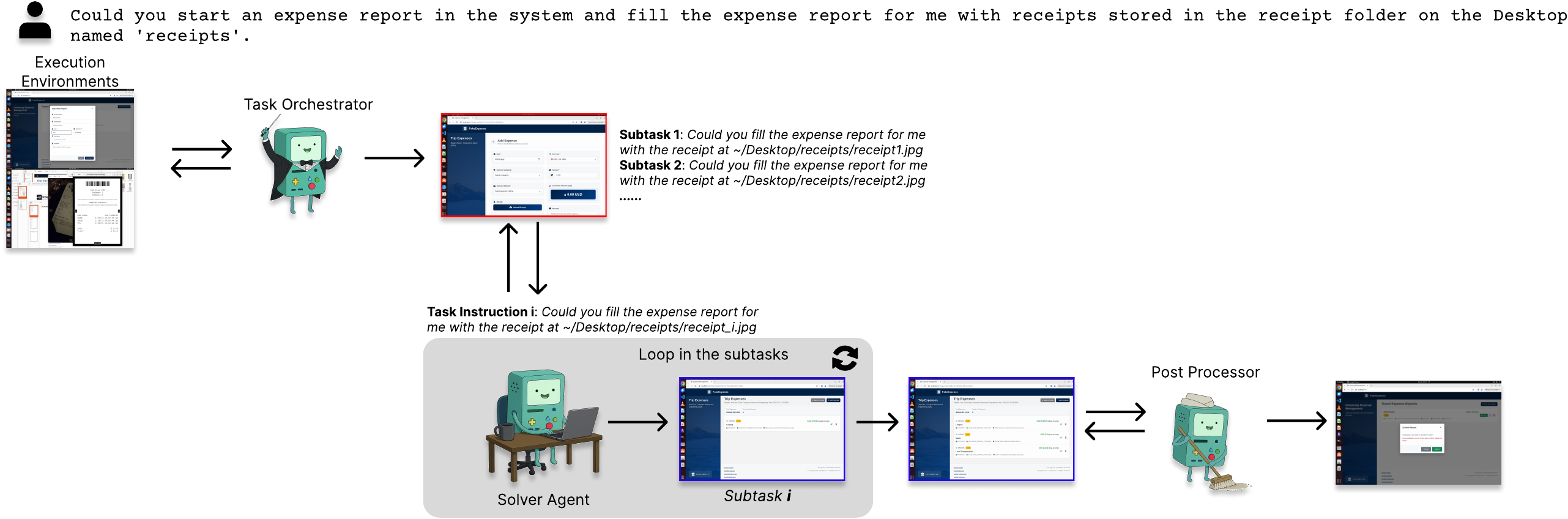}
    \caption{\textbf{Orchestrator framework}. The orchestrator handles the pre-subtask phase, a set of solver agents executes the repetitive phase, and the post-processor handles the post-subtask phase.}
    \label{app:orchestrator-framework}
\end{figure*}

We adapt the orchestrator idea from \cite{Wang2025OdysseyBenchEL}; the framework is illustrated in \Cref{app:orchestrator-framework}. 
The framework adopts multiple agents for distinct roles, one per phase of the three-phase task structure defined in \Cref{sec:task-definition}:
A \textbf{task orchestrator} $\mathbf{O}$ handles the pre-subtask phase: it explores the environment, parses the user intention, fills in the shared meta-information (e.g., the employee's name and ID before starting an expense report), and decomposes the workflow into per-instance subtask instructions.
A set of \textbf{solver agents} $\{\mathbf{S}_i\}_{i=1}^{N}$ executes the repetitive subtask phase. 
The framework iterates over the $N$ data instances in a for-loop, initialising a solver agent with the corresponding subtask instruction at each iteration and moving on once the agent reports subtask completion.
A \textbf{post-processor} $\mathbf{P}$ handles the post-subtask phase once all subtasks are completed. 
This framework theoretically decomposes a vast-horizon, repetitive task into shorter, more tractable subtasks.
All three roles, i.e. orchestrator $\mathbf{O}$, solver $\mathbf{S}_i$, and post-processor $\mathbf{P}$, are instantiated as separate AgentS2.5 + GPT-5 agents.

\begin{table*}[!t]
\centering
\tiny
\resizebox{0.8\linewidth}{!}{
\begin{tabular}{l|cc}
\toprule
\bfseries Agents        & \multicolumn{1}{l}{\bfseries Field-Level SR} & \multicolumn{1}{l}{\bfseries Subtask-Level SR} \\
\midrule
Agent S2.5 + GPT5                               & 33.90                               & 25.31                                \\
Agent S2.5 + GPT5 + \ourmethod              & 50.95                              & 41.24                                \\
\midrule
Agent S2.5 + GPT5 + Orchestrator                & 29.07                              & 16.01                                \\
Agent S2.5 + GPT5 + Orchestrator + \ourmethod & 40.78                              & 27.40                                 \\
\midrule
GPT5.4-Mini                                     & 09.90                                & 07.35                                 \\
GPT5.4-Mini + \ourmethod                      & 27.34                              & 18.66   \\
\bottomrule
\end{tabular}
}
\caption{\textbf{\ourmethod provides consistent gains across different agent configurations.} Applied to three different setups (AgentS2.5 + GPT-5, AgentS2.5 + GPT-5 + Orchestrator, and GPT5.4-mini), \ourmethod improves both field-level and subtask-level success rates in every setting, disentangling the method's contribution from the underlying agent's capacity. Each cell reports the success rate in percentage (\%).}
\label{app:method-ablation}
\end{table*}

\paragraph{Adapt \ourmethod to the orchestrator framework.}
The graph-based demonstration is partitioned into three sub-graphs along the orchestrator's three-phase decomposition, one for the task orchestrator $\mathbf{O}$, one for the solver agents $\mathbf{S}_i$, and one for the post-processor $\mathbf{P}$, so each component receives only the demonstration content relevant to its role.
Each subagent is initialised with the text-level abstraction of its own sub-graph in the system prompt.
The rest of the demonstration segment retrieval, context embedding stays the same. 

\subsubsection{More details about \ourmethod}
\label[appendix]{app:method-details}

\paragraph{Implementation details. }
We further introduce the implementation details of \ourmethod. 
Nodes are organised hierarchically: root nodes capture higher-level semantic stages (e.g., \textbf{\texttt{navigate\_to\_the\_folder}}), while child nodes record the step-by-step trajectory within each stage (e.g. \textbf{\texttt{navigate\_to\_Desktop}}, \textbf{\texttt{navigate\_to\_receipts\_folder}}, etc.). 
Directed edges carry the \textbf{\texttt{action}} executed between states. 
Each leaf node additionally stores \textbf{\texttt{action\_reason}} and \textbf{\texttt{action\_goal}} for context of the current action. 
This hierarchical organisation groups the demonstration into coherent semantic units, enabling more efficient retrieval.




\paragraph{More quantitative analysis of \ourmethod.}
To disentangle \ourmethod's benefit from agent capacity, we extend \Cref{tab:orchestrator} with results on the weaker GPT5.4-Mini.
\Cref{app:method-ablation} reports field-level and subtask-level success rates for each configuration with and without \ourmethod. Across all three settings, \ourmethod consistently improves the performance, indicating that the gains do not come from any single agent's underlying capacity.





\subsection{More Details About \ourbenchmark}
\begin{figure*}
    \centering
  \includegraphics[width=\linewidth]{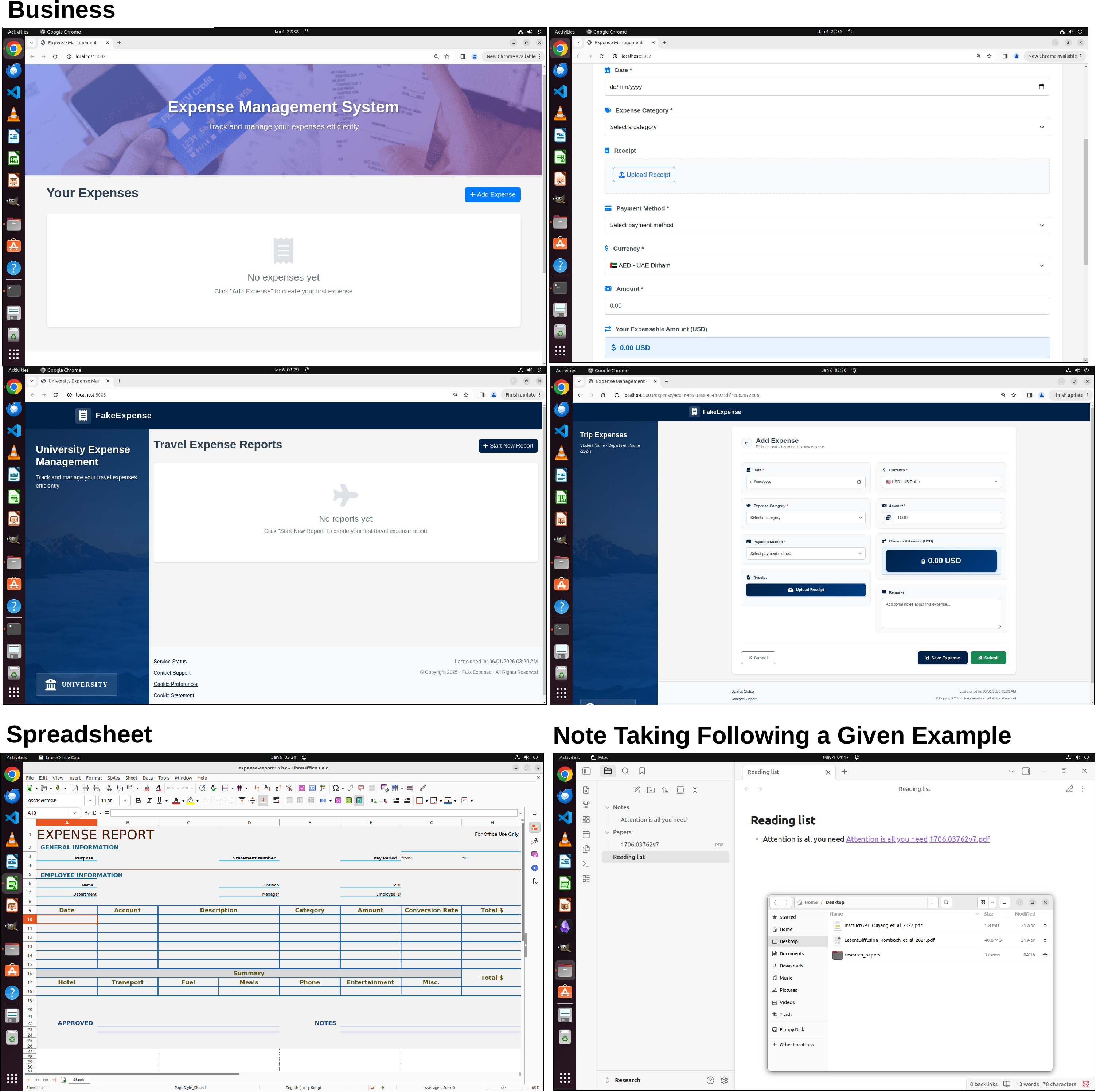}
  \caption{Examples of the execution environment in the business and note taking scenarios.}
  \label{fig:appen-exec-env1}
\end{figure*}

\begin{figure*}
    \centering
  \includegraphics[width=\linewidth]{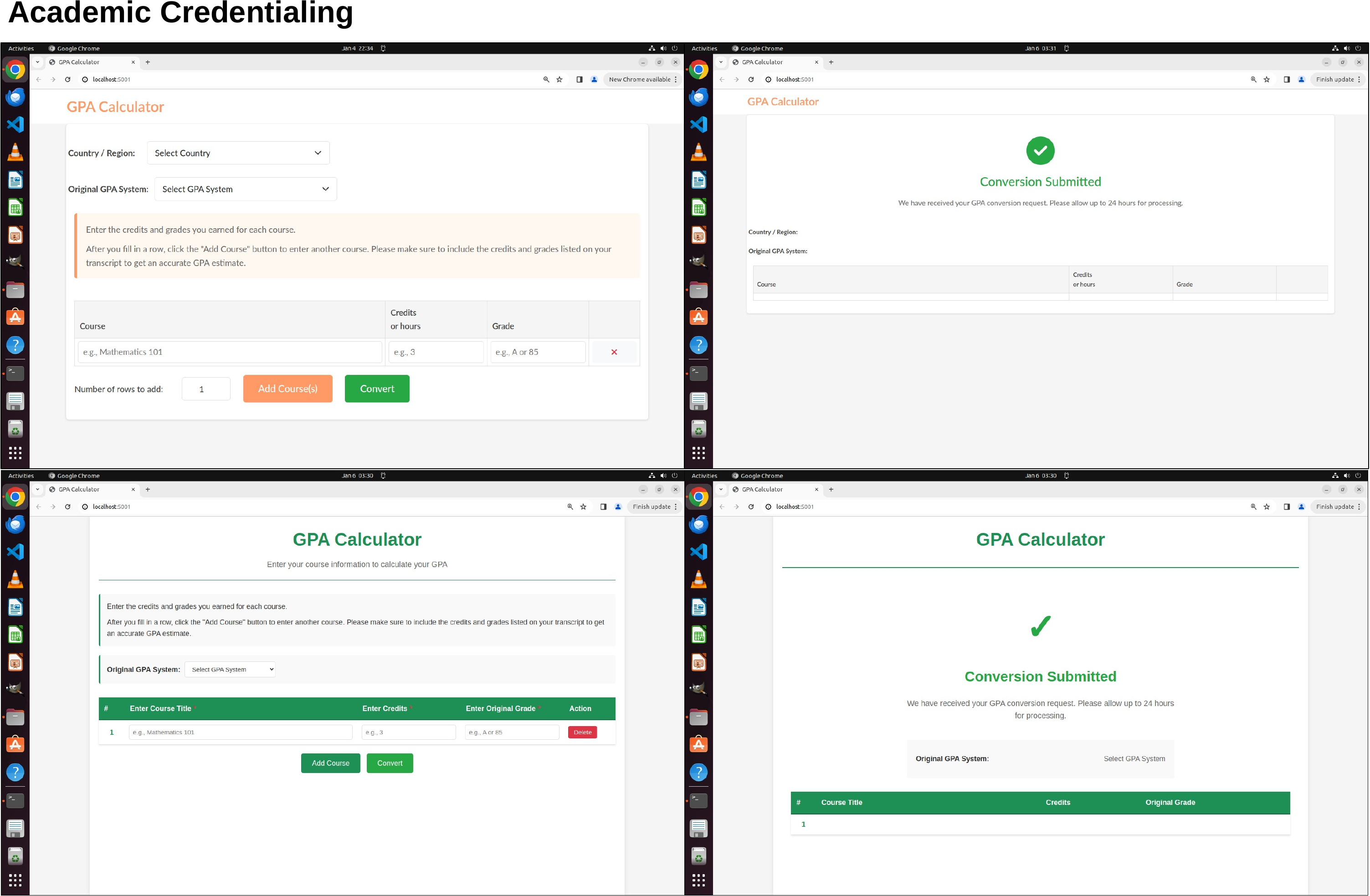}
  \caption{Examples of the execution environment in the academic credentialing scenario.}
  \label{fig:appen-exec-env2}
\end{figure*}

\begin{figure*}
    \centering
  \includegraphics[width=\linewidth]{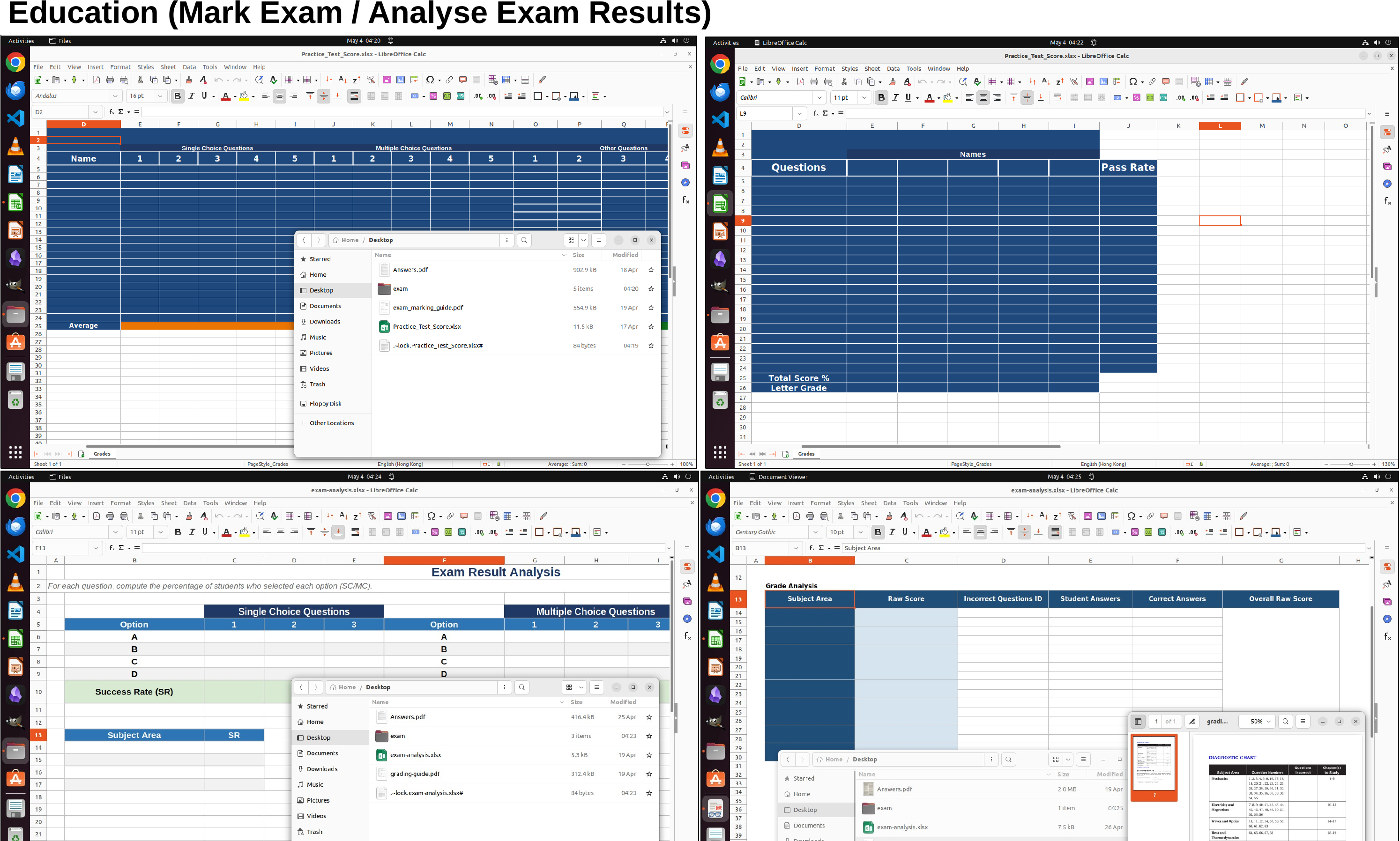}
  \caption{Examples of the execution environment in the education scenario.}
  \label{fig:appen-exec-env3}
\end{figure*}

\begin{figure*}
    \centering
  \includegraphics[width=\linewidth]{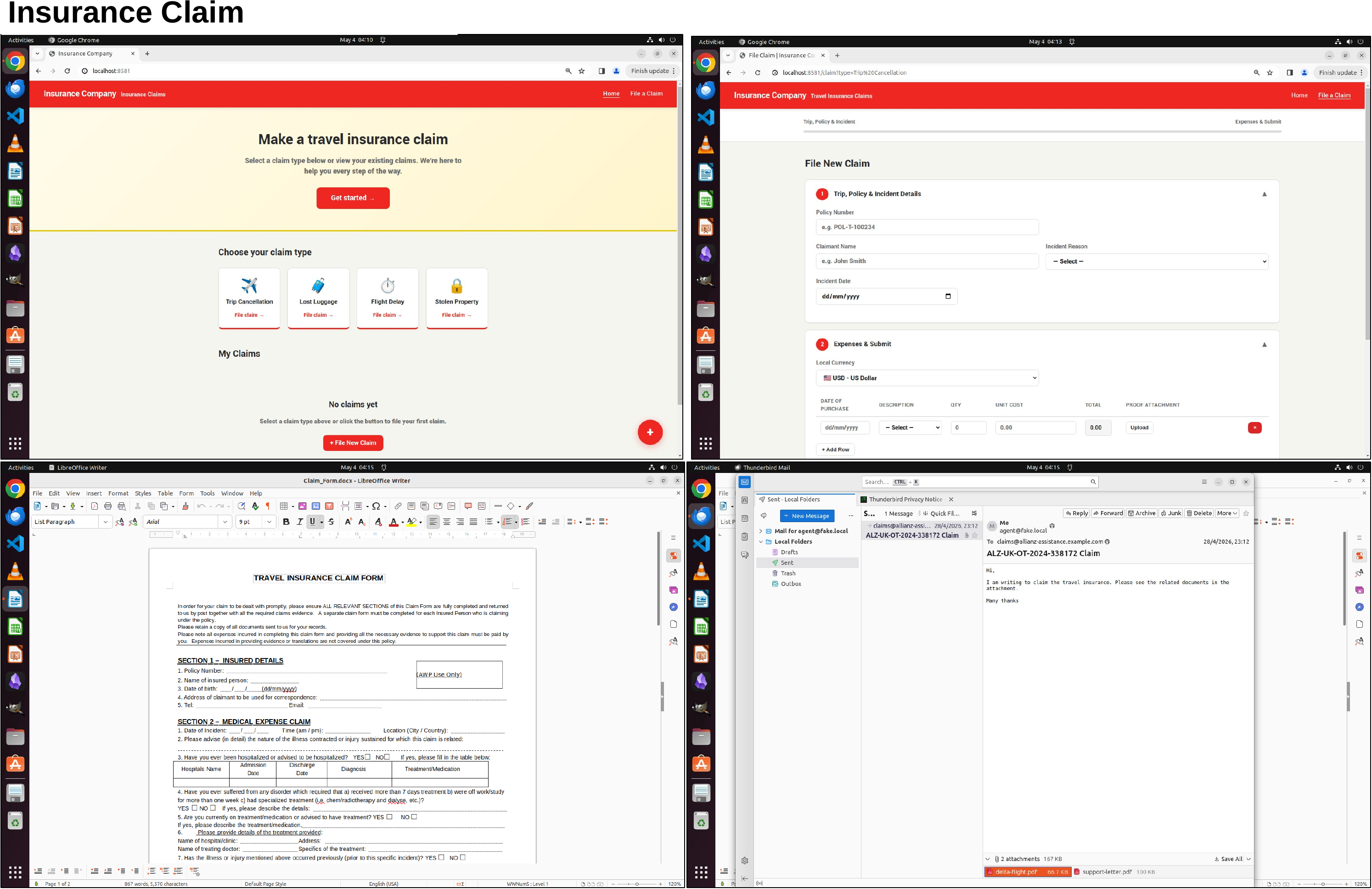}
  \caption{Examples of the execution environment in the insurance scenario.}
  \label{fig:appen-exec-env4}
\end{figure*}

\paragraph{Execution Environment Examples. }
\label[appendix]{app:appen-exeuction-example}
\Cref{fig:appen-exec-env1,fig:appen-exec-env2,fig:appen-exec-env3,fig:appen-exec-env4} visualise the all the execution environments included in \ourbenchmark. 



\end{document}